\documentclass[twoside]{article}

\usepackage[accepted]{aistats2026}

\usepackage[round]{natbib}

\usepackage{amsmath,amsthm,amsfonts,amssymb,mathtools}
\usepackage{enumitem}
\usepackage{bm,dsfont}
\usepackage[T1]{fontenc}
\usepackage{xcolor,pifont}
\usepackage{graphicx}
\usepackage{caption, subcaption, float}
\usepackage{hyperref}
\usepackage[nameinlink]{cleveref}
\usepackage[title]{appendix}
\usepackage{algorithm,algorithmic,setspace}
\usepackage{booktabs}
\usepackage{multirow}
\usepackage{makecell}
\usepackage{color, colortbl, tcolorbox}
\usepackage[hang,flushmargin]{footmisc}
\usepackage{ifthen}

\newtheorem{theorem}{Theorem}

\definecolor{Gray}{gray}{0.9}
\usepackage[textsize=tiny]{todonotes}
\allowdisplaybreaks

\begin{document}

\runningauthor{Zhou, Tian, and Diggavi}
\runningtitle{An IT-Approach to Understanding Transformers' ICL of VOMCs}
\twocolumn[
\aistatstitle{An Information-Theoretic Approach to Understanding Transformers' In-Context Learning of Variable-Order Markov Chains}
\aistatsauthor{Ruida Zhou  \And Chao Tian \\ \And  Suhas Diggavi}
\aistatsaddress{Amazon AGI \And ECE Dept. 
   Texas A\&M University  \And EE Dept.  UCLA }]
\makeatletter
\pagestyle{fancy}
\fancyhead[CE]{\small\bfseries\@runningtitle}
\fancyhead[CO]{\small\bfseries\@runningauthor}
\makeatother

\begin{abstract}
We study transformers' in-context learning of variable-length Markov chains (VOMCs), focusing on the finite-sample accuracy as the number of in-context examples increases. Compared to fixed-order Markov chains (FOMCs), learning VOMCs is substantially more challenging due to the additional structural learning component. The problem is naturally suited to a Bayesian formulation, where the context-tree weighting (CTW) algorithm, originally developed in the information theory community for universal data compression, provides an optimal solution. Empirically, we find that single-layer transformers fail to learn VOMCs in context, whereas transformers with two or more layers can succeed, with additional layers yielding modest but noticeable improvements. In contrast to prior results on FOMCs, attention-only networks appear insufficient for VOMCs. To explain these findings, we provide explicit transformer constructions: one with $D+2$ layers that can exactly implement CTW for VOMCs of maximum order $D$, and a simplified two-layer construction that uses partial information for approximate blending, shedding light on why two-layer transformers can perform well.
\end{abstract}

\vspace{-0.1cm}
\section{INTRODUCTION}
\vspace{-0.1cm}
The transformer model \citep{vaswani2017attention}, the architecture behind current prevailing LLMs, is known to have strong in-context learning (ICL) capabilities, and concrete ICL results for transformers have been established for some simple tasks \citep{garg2022can,von2023transformers,bai2024transformers,ahn2024transformers,zhang2025training}. Despite these advances, the mechanism for transformers to learn in context is still not fully understood, especially when the sequences have complex memory structures. 
Several recent works studied how transformers can learn fixed-order Markov chains (FOMCs) either in training or in-context \citep{makkuva2024attention,edelman2024evolution,rajaraman2024transformers,nichani2024transformers,akyurek2024context,zhang2025what}, however, FOMC has a pre-defined fixed memory structure, which is highly inefficient in modeling complex memories. In this work, we study the ICL of a considerably more complex model, i.e., variable-order Markov chains (VOMCs), also known as context tree (CT) models \citep{rissanen1983universal,willems1995context},  which is better suited for sequences with more complex memory structures such as natural languages \citep{begleiter2004prediction}; we refer to the problem as ICL-VOMC. 

Although VOMCs still have finite memory, each new symbol may depend on suffixes consisting of different numbers of previous symbols, i.e., the length of the memory may vary. For example, in the sentence below, 
\begin{center}
\vspace{-0.1cm}
\fbox{\begin{minipage}{0.46\textwidth}
Language models are \underline{useful in a wide} \textcolor{blue}{variety} of applications, \underline{from natural language} \underline{understanding and generation to translation,} \textcolor{blue}{summarization}, and...
\end{minipage}}
\vspace{-0.1cm}
\end{center}
``variety'' and ``summarization'' can be modeled as depending naturally (only) on their respective underlined suffixes of different lengths. The inherent structural learning component makes it considerably more challenging, and the naive strategy of adopting an FOMC-based approach is highly sample-inefficient. 

Instead of pretraining dynamics, we focus on the ICL accuracy as more in-context examples are given, i.e., the finite in-context sample performance. 
The well-known context-tree weighting algorithm (CTW) \citep{willems1995context}, originally developed in the information theory community for universal data compression, is in fact Bayesian optimal for the in-context inference task. This is because the underlying core component of the CTW algorithm estimates the next token probability distribution -- essentially the same task the transformers perform for ICL. \cite{kontoyiannis2022bayesian} provided a more detailed analysis of the algorithm from a Bayesian inference perspective; earlier work along the Bayesian view can be found in \cite{matsushima1994bayes,matsushima2002class}. The connection between ICL and data compression is not entirely surprising, given the recent empirical study \citep{deletang2023language} showing LLMs' superior performance on various compression tasks.

We empirically observed that one-layer transformers cannot learn VOMCs, but those with two or more layers can learn VOMCs and track the performance of the optimal CTW algorithm closely, with more layers providing small but noticeable improvements. Moreover, attention-only networks suffer materially. We then consider explicit transformer constructions to explain these findings. The CTW algorithm has a recursive structure, which poses a challenge in the transformers' condensed parallel computation pipeline. To resolve this, we first identify an alternative representation of the CTW algorithm, and then construct a transformer with $D+2$ layers, which can learn VOMC of maximum order $D$ in context. The construction relies on the feedforward layers, which explain why attention-only networks perform poorly on VOMCs.

The CTW algorithm needs the counts for the suffixes, and the higher layers of our transformer construction are mostly performing information aggregation. We conjecture that 2-layer transformers can perform reasonably well because even raw counts or partially aggregated information would allow a close-to-optimal approximation. We then consider such a simplified 2-layer transformer, by providing one feed-forward (FF) layer with the probability estimates and various configurations of the counts directly. The FF layer can be trained to approximate the proper blending coefficients using this information. We implement several synthetic transformer layers, and experiments show that their performances indeed match our conjectures. 

It may be tempting to directly invoke the universal computation capability of transformers, e.g. \cite{perez2021attention}, without seeking explicit interpretations, however, these results usually do not provide any transparency on the underlying mechanisms. In contrast, we focus on constructions that provide meaningful \textit{interpretations} of the empirical observations, with a focus on the mechanisms that extract and aggregate the statistics (more discussions after Theorem \ref{thm:induction}). We further remark that the VOMC model provides a complex but clean setting where the optimal ICL performance is known and other confounding factors can be well-controlled, which is impossible with real-world language datasets and benchmarks. 

\paragraph{Main Contributions.} Our work is the first study of ICL-VOMC, particularly the {\em finite-sample performance}, and the contributions are summarized below. 
\begin{enumerate}[itemsep=-0.5pt,topsep=-1pt]
\item We empirically demonstrate that transformers can learn VOMCs in-context, i.e., tracking the performance of the optimal CTW algorithm in the whole context window;
\item We provide a $(D+2)$-layer transformer construction to imitate CTW, establishing its capabilities, based on a novel Bayesian optimal next token prediction representation;
\item Our construction allows us to investigate the relative performance insensitivity to the number of layers, \emph{i.e.,} why $2$-layer transformers perform well. 
\end{enumerate}

\textbf{Notation:} Scalars, symbols, and strings are denoted by italic letters like $n, N$, $x$, and $s$. Denote by string $x_{i}^{j} := (x_{i}, x_{i+1},\ldots,x_{j})$ as a sequence of symbols. Define $()$ or $x_{i}^{j}$ with $i>j$ as an empty string. Vectors and matrices are in bold like $\vec{x}$, $\vec{H}$, and sets in calligraphic like $\mathcal{A}$ with cardinally $|\mathcal{A}|$.

\vspace{-0.1cm}
\section{PRELIMINARIES}
\label{sec:pre}

\vspace{-0.1cm}

\paragraph{Universal Data Compression.} ICL closely resembles universal compression \citep{rissanen1983universal} in information theory. Na\"ively speaking, the latter aims to adaptively compress the sequence without having direct access to the underlying probabilistic dynamics, but learning it in an in-context fashion. This implies that the CTW algorithm, though originally designed for data compression, is Bayesian optimal for ICL-VOMC. One subtle point is that the {\em finite-sample} in-context performance is our main interest, i.e., transformers are viewed as performing ICL-VOMC efficiently, only if the performance over the whole context window can track that of CTW; asymptotical (large in-context sample) matching performance is insufficient. 

\vspace{-0.1cm}
\paragraph{Variable-order Markov Chains.} VOMCs have been studied extensively in the information theory literature \citep{rissanen1983universal,willems1995context,begleiter2004prediction}. String $s=(x_{1-l},x_{2-l},\ldots,x_{0})$ is a suffix of the string $s'=(x'_{1-l'},x'_{2-l'},\ldots,x'_{0})$, if $0\leq l\leq l'$ and $x_{-i}=x'_{-i}$ for $i=0,1,\ldots,l-1$; e.g., $(a,b,c,b)$ is suffix of $(a,c,a,a,b,c,b)$. Note that the strings above have non-positive indices. 

\begin{figure}[t!]
    \centering    
        \includegraphics[width=0.42\textwidth]{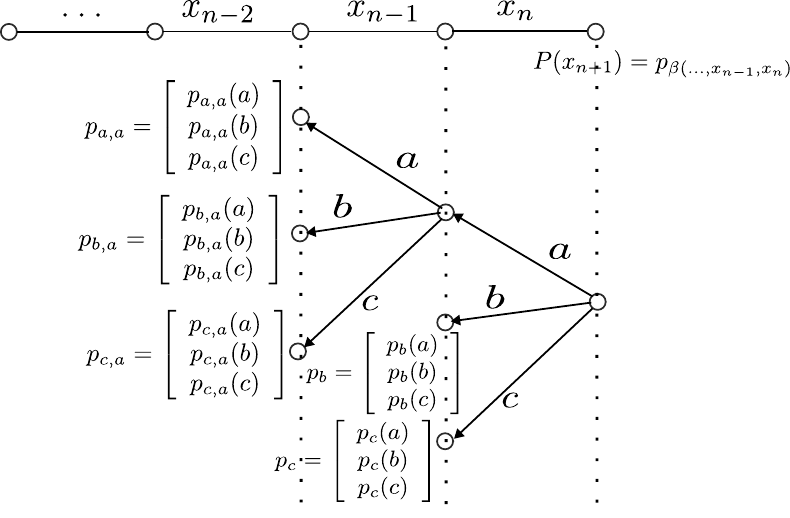}
        \caption{ A context tree in the alphabet $\mathcal{A}=\{a,b,c\}$ with suffix set $\mathcal{S}=\{(b),(c),(a,a),(b,a),(c,a)\}$ and the  probability distributions. If $(\ldots,x_{n-1},x_n)=(\ldots,c,a)$, then the probability distribution for the next symbol $x_{n+1}$ is $p_{c,a}$.\label{fig:CT}}    
\vspace{-0.3cm}
\end{figure}

The behavior of a finite memory VOMC source is specified by a suffix set $\mathcal{S}$ and the associated next token probability distributions. The set $\mathcal{S}$ is a collection of strings $s(k)$, $k=1,2,\ldots,|\mathcal{S}|$ with two properties: 1) proper: no string in $\mathcal{S}$ is a suffix of any other string, and 2) complete: each semi-infinite sequence $(\ldots,x_{n-1},x_{n})$ has a unique suffix that belongs to $\mathcal{S}$, denoted as $\beta_{\mathcal{S}}(\ldots,x_{n-1},x_{n})$. Associated with each suffix $s\in\mathcal{S}$, there is a probability mass function $p_s\in \Delta_{\mathcal{A}}$. A VOMC is best visualized as a context tree, since for any valid suffix set $\mathcal{S}$, there exists a unique tree $T$ with the elements in $\mathcal{S}$ being its leaves; the set of leaves is denoted as $\mathcal{L}(T)$. A CT can thus be equivalently represented by $(T, \{p_s\}_{s \in \mathcal{L}(T)})$; an example is given in Fig. \ref{fig:CT}. A VOMC has a maximum order $D$ if any suffix in $\mathcal{S}$ has a length at most $D$. Given a semi-infinite sequence $(\ldots,x_{n-1},x_n)$, the next symbol $x_{n+1}$ is generated according to the distribution $p_{\beta_{\mathcal{S}}(\ldots,x_{n-1},x_{n})}$. 

\begin{figure}[t!]
  \centering
   \includegraphics[width=0.395\textwidth]{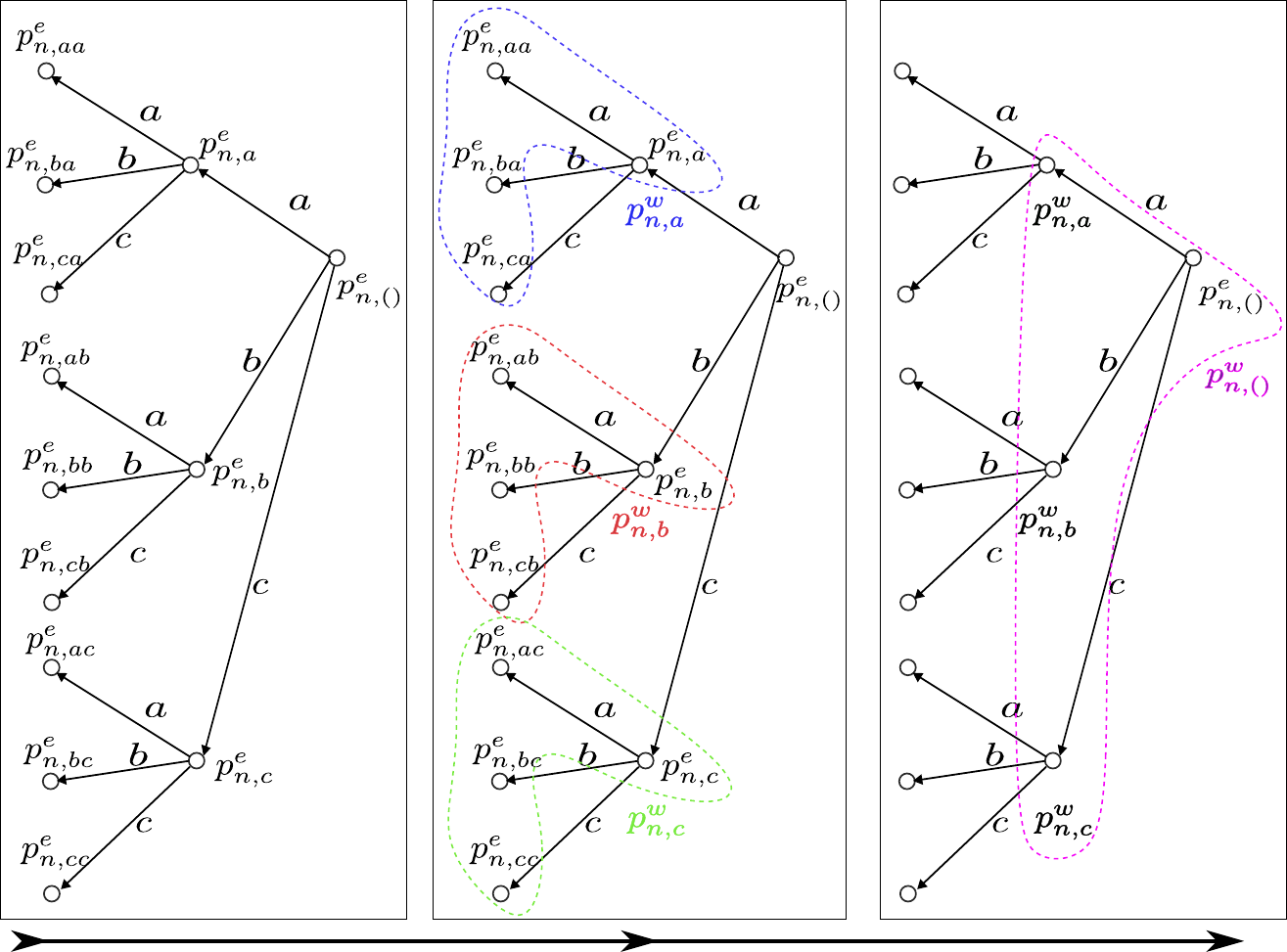}   
    \caption{The CTW algorithm aggregates statistics from the leaves to the root in a recursive manner at time position $n$ (here $D=2$).\label{fig:main}}    
    \vspace{-0.3cm}
\end{figure}

The CTW algorithm is Bayesian optimal for VOMCs under certain priors, i.e., achieving the optimal cross-entropy loss. An illustration of the algorithm is given in Fig. \ref{fig:main} (the details are deferred to Section \ref{sec:CTW}). Prediction by partial matching (PPM) \citep{cleary1984data,begleiter2004prediction} and Kneser-Ney (KN) smoothing \cite{ney1994structuring} can also be used to generate the next token prediction. They take the maximal order $D_{\text{ppm}}$ or $D_{\text{kn}}$ as parameters, and when a suffix has not been observed at all (or is less often observed), the prediction falls back to the estimate using a lower-order model. PPM uses an escape symbol to indicate the fallback, while KN-smoothing uses a soft blending instead. Both can be viewed as based on FOMC modeling, since they mostly adopt the prediction at the order $D_{\text{ppm}}$ or $D_{\text{kn}}$ when possible. More details are given in Appendix \ref{app:ppm} and \ref{app:kn}.

\vspace{-0.1cm}
\section{IN-CONTEXT LEARNING OF VOMCS}
\begin{figure}[t!]
\centering
\includegraphics[width=0.44\textwidth,trim={1cm 0.4cm 1cm 1cm},clip]{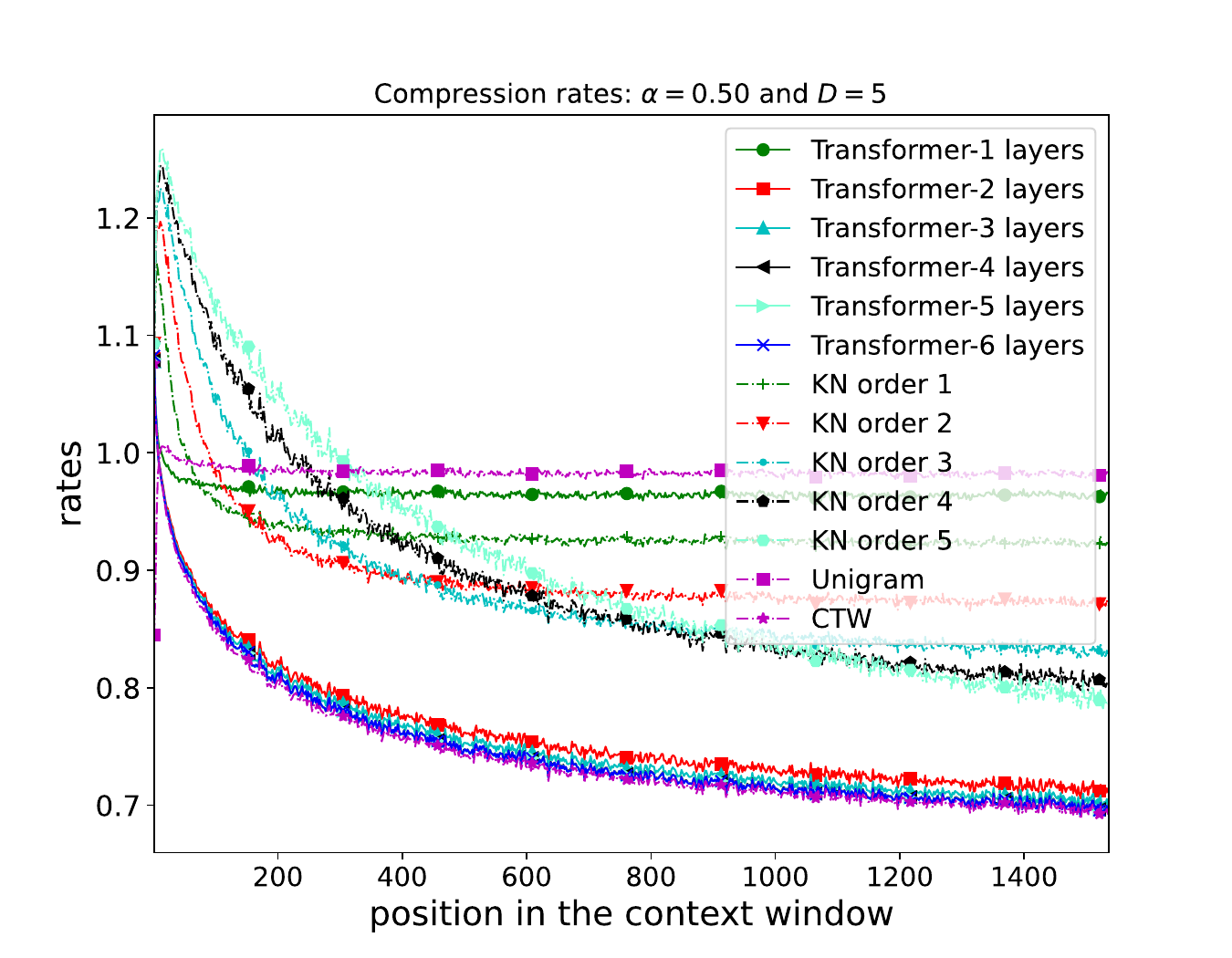}
\vspace{-0.1cm}
\caption{Transformers vs. KN-smoothing vs. CTW. \label{fig:tf5}}
\vspace{-0.3cm}
\end{figure}

\vspace{-0.1cm}
\paragraph{Transformers Can Learn VOMC In-context.} We train transformers to conduct experiments. A ternary alphabet $|\mathcal{A}|=3$ is adopted, and transformers of context window size $N$ are pretrained on data sequences of length-$N$ generated using CTs randomly sampled from the given prior $\pi_{\text{CTW}}$ parameterized by $\boldsymbol{\alpha}=0.5$, $\lambda=0.15$ and a maximum tree depth $D$; training details are given in Appendix \ref{app:pretrain-details}. The training loss is the sum of canonical cross-entropy losses in the context window, or equivalently, the sum of data compression rates. During testing, given a source sequence of length-$N$ generated from an unknown VOMC with an order at most $D$, can the transformer learn this sequence efficiently, i.e., at a loss close to the optimal value over the whole context window? In Fig. \ref{fig:tf5}, we show the ICL performances (the compression rates, equivalent to the cross-entropy losses) of the trained transformers with various numbers of layers, and the CTW and KN algorithms for $N=1536$ and CT maximum order $D=5$. The transformers have 8 attention heads with embedding dimension $E=128$. The horizontal axis is the position in the context window, and the rates indicate how well transformers and other algorithms learn the underlying VOMCs in context.

\begin{figure}[t!]
\centering
\includegraphics[width=0.43\textwidth,trim={1.05cm 0.4cm 1cm 1cm},clip]{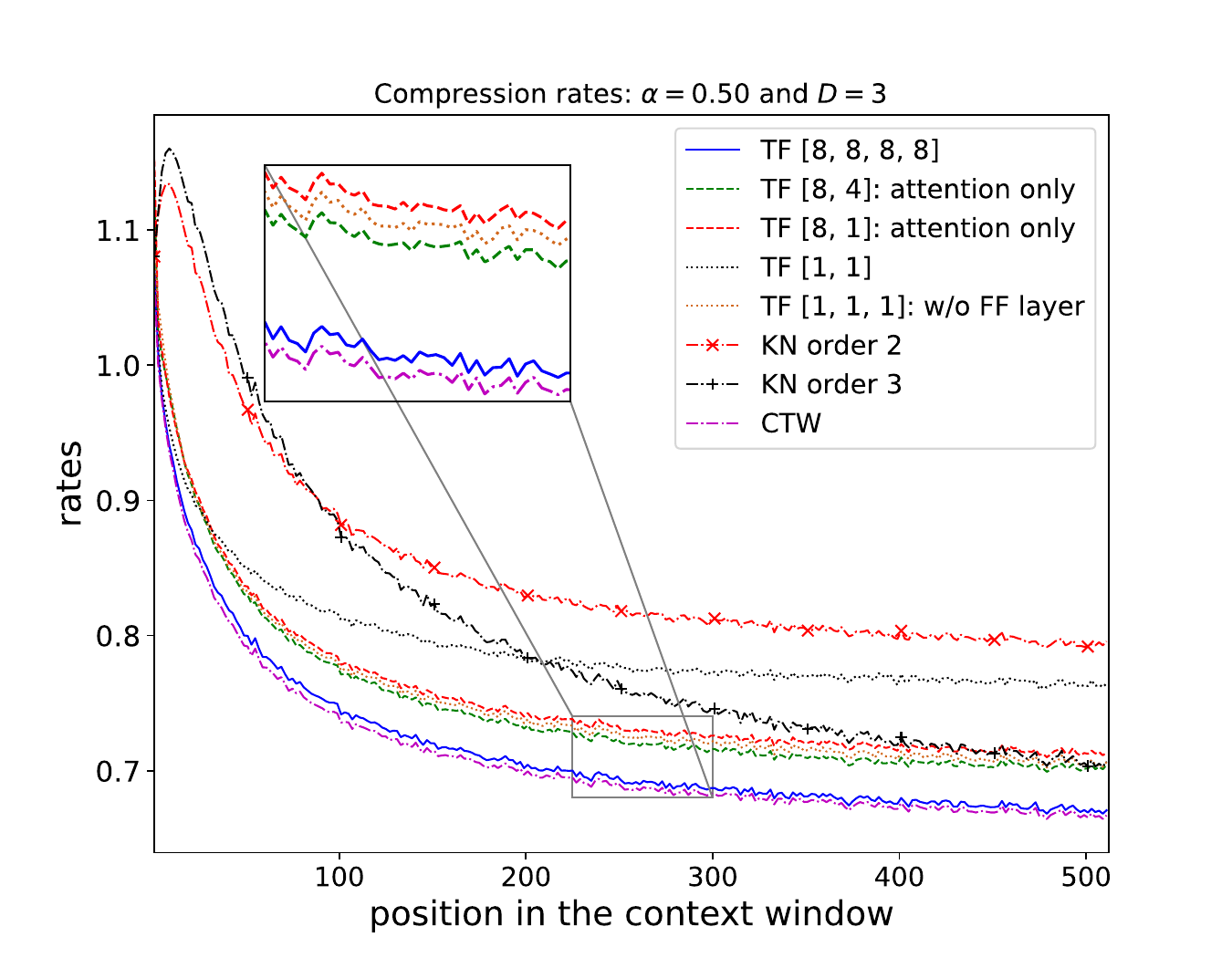}
\vspace{-0.1cm}
\caption{Attention-only vs. full transformers. \label{fig:attn}}
\vspace{-0.3cm}
\end{figure}

CTW is Bayesian optimal in this setting, whereas the KN-smoothing (with optimized hyperparameter) performs poorly, which is expected since it essentially reduces to an FOMC estimator at the assumed maximum order $D_{\text{kn}}$. If $D_{\text{kn}}<D$, then there is oversimplification; even when $D_{\text{kn}}$ is sufficiently large, i.e., $D_{\text{kn}} \geq D$, it is a highly inefficient estimator for those suffixes at lower orders, thus performing poorly in finite-length context window. This effect is particularly pronounced at the latter part of the context window, e.g., a loss floor manifests when $D_{\text{kn}}<D$; at the beginning part of the context window, the potentially inaccurate prior and ill-matched FOMCS updates (escape symbols) jointly make KN's performance poorer. We can view KN as a reference method that does not adapt to variable orders. Additional experimental results with PPM are given in Appendix \ref{app:additional-expriments}.

Almost all trained transformers, except those with a single layer, can track the performance of the CTW algorithm fairly closely in the context window. The single-layer transformer's performance is similar to (only slightly better than) the unigram approach. Transformers' overall performance improves as the number of layers increases in general. Nevertheless, the improvements with increased numbers of layers are relatively small and appear to saturate at four layers. A construction is given in Section \ref{sec:attention-map} to explain this phenomenon. We further notice that even transformers with two layers appear to learn in context quite well. In Section \ref{sec:why2layer}, we provide an explanation of this phenomenon using a simplified transformer construction and additional experimental results.

\vspace{-0.1cm}
\paragraph{Insufficiency of Attention-only Networks.}  Previous works \cite{edelman2024evolution,nichani2024transformers} showed that a two-layer attention-only network (i.e., without attention layer normalization or the FF layer) can effectively perform ICL for FOMCs, where the second layer only requires a single attention head. Moreover, the authors of \cite{rajaraman2024transformers} empirically observed that two-layer single-head transformers can perform ICL-FOMCs of at least order 4, and constructed a three-layer single-head attention-only network (without an FF layer but with attention layer normalization) that can perform ICL-FOMCs. We conducted experiments to verify whether these continue to hold for ICL-VOMCs, and the results are shown in Fig. \ref{fig:attn}, where the bracket after TF (transformer) indicates the number of heads in each layer. Observe that attention-only networks perform considerably worse than CTW and a full four-layer transformer. Two-layer single-head transformers, and three-layer single-head transformers without the FF layer, also perform poorly, particularly the former. These results show that the mechanisms given in \cite{edelman2024evolution,nichani2024transformers,rajaraman2024transformers} for ICL-FOMCs cannot explain well how transformers perform ICL-VOMC, and the feedforward networks need to play a more significant role. 

\vspace{-0.1cm}
\section{THEORETICAL INTERPRETATIONS}
\label{subsec:CT}
\vspace{-0.2cm}
\subsection{The Transformer Architecture} \label{sec:transformer}
\vspace{-0.1cm}

We introduce a mathematical description of the $L$-layer decoder-only transformer model to fix the notation. Transformer interacts with sequential data, e.g., $x_1^N = (x_1, \ldots, x_{N})$, where token $x_i$ is a symbol from an alphabet (a.k.a. vocabulary) $\mathcal{A}$ with $A = |\mathcal{A}|$. Each token $x_i$ is embedded into $\vec{h}^{(1)}_i \in \mathbb{R}^{E}$ by integrating the information of its value $x_i$ and position $i$, where $E$ is the embedding dimension. 

 Each layer of the transformer takes matrix $\vec{H}^{(\ell)}=[\vec{h}^{(\ell)}_1,\vec{h}^{(\ell)}_2,\ldots,\vec{h}^{(\ell)}_N]$, where $\vec{h}_i^{(\ell)}\in \mathbb{R}^E$, as its input and applies the multi-head attention (MHA) layer operation and the feed-forward layer operation, and the output is the input to the next layer, denoted as $\vec{H}^{(\ell+1)}$. The output of the  multi-head attention (sub-)layer with $M^{(\ell)}$ heads is
\begin{align*}
\vec{a}_{i}^{(\ell)}&=\text{MHA}\left(\vec{h_i},\vec{H};\{W_{O,m}^{(\ell)},W_{Q,m}^{(\ell)},W_{K,m}^{(\ell)},W_{V,m}^{(\ell)}\}_{m=1}^{M^{(\ell)}}\right)\\
&\triangleq W_O^{(\ell)}\left[\vec{b}_{1,i}^{(\ell)};\vec{b}_{2,i}^{(\ell)};\ldots;\vec{b}_{M^{(\ell)},i}^{(\ell)}\right],
\end{align*}
where $\{W_{Q,m}^{(\ell)},W_{K,m}^{(\ell)},W_{V,m}^{(\ell)}\}_{m=1}^{M^{(\ell)}}$ are the $E^{(\ell)} \times E$ query matrices, key matrices, and value matrices\footnote{In practice, embedding dimension $E$ is divisible by the number of heads $M^{(\ell)}$ and $E = M^{(\ell)} E^{(\ell)}$.} at the $\ell$-th layer and $m$ is the index of the attention head, respectively, $W_{O}^{(\ell)}$ is the $E \times (M^{(\ell)}E^{(\ell)})$ output mapping matrix,  and $\vec{b}_m^{(\ell)}$ is the output of the $m$-th attention head at this layer defined as
\begin{align*}
&\vec{b}_{m,i}^{(\ell)} = (W_{V,m}^{(\ell)}[\vec{h}_1^{(\ell)},\vec{h}_2^{(\ell)},\ldots,\vec{h}_i^{(\ell)}])\\
&\quad\cdot\text{softmax}((W^{(\ell)}_{K,m} [\vec{h}^{(\ell)}_1,\vec{h}^{(\ell)}_2,\ldots,\vec{h}^{(\ell)}_i])^\top(W^{(\ell)}_{Q,m} \vec{h}^{(\ell)}_i)), %\label{eqn:attention}
\end{align*}
where we used ``$;$'' to indicate vertical matrix concatenation and ``$,$'' to indicate horizontal matrix concatenation. The attention layer has a residual connection, and the attention output together with the residual connection also goes through a feed-forward (sub-)layer with a residual connection
\begin{align*}
\vec{h}_i^{(\ell+1)}&=\text{FF}(\vec{a}_i^{(\ell)} ;W^{(\ell)}_1,W^{(\ell)}_2)\\
&= W^{(\ell)}_1\sigma(W^{(\ell)}_2 (\vec{a}^{(\ell)}_i+\vec{h}_i^{(\ell)}))+(\vec{a}^{(\ell)}_i+\vec{h}^{(\ell)}_i),
\end{align*}
where $\sigma$ is a non-linear activation function (e.g., ReLU or sigmoid). The output of the last ($L$-th) transformer layer $\vec{H}^{(L+1)}$ goes through a linear then softmax to predict the probability of generating the next symbol in the vocabulary $\mathcal{A}$ based on the past observations: 
\begin{align*}
&\hat{\vec{p}}_{i+1} = \text{softmax}(W^{(L+1)}_O \vec{h}_i^{(L+1)}) \in \Delta_{A},\notag\\
&\qquad\qquad\qquad\qquad\qquad\,i=1,\ldots,N-1, 
\end{align*}
where $\Delta_{A}$ is the probability simplex on $\mathcal{A}$. The model is illustrated in Appendix \ref{app:transformer}.

Transformers are (pre)-trained to predict next token by minimizing the cumulative log-loss (cross-entropy loss) $\mathbb{E}_{x_{1}^{N}}[ \sum_{i = 1}^{N-1} \vec{x}_{i+1}^\top \log(1/\hat{\vec{p}}_{i+1})]$, where $\vec{x}_i \in \mathbb{R}^{A}$ is the one-hot encoding of $x_i$ and sequence $x_{1}^N$ is sampled from some population of sources, e.g., sequences can be articles written by different authors and thus following different dynamics.

\vspace{-0.1cm}
\subsection{The Bayesian CTW Algorithm}
\label{sec:CTW}
\vspace{-0.1cm}

The difficulty to identify and estimate accurately the underlying CT is in learning both of its components: the tree structure itself, and the probability distribution associated with each leaf node. The likelihood of a sequence $x_1^i$ given $x_{1-D}^0$ for a CT with parameter $(T, \{p_s\}_{s \in \mathcal{L}(T)} )$ is 
\vspace{-0.1cm}
\begin{align}
    P_{ T, \{p_s\} }( x_1^i | x_{1-D}^0 ) &= \prod_{j = 1}^i p_{\beta_{\mathcal{L}(T)}(x_{j-D}, \ldots, x_{j-1})}(x_j)\notag\\
    &= \prod_{s \in \mathcal{L}(T)} \prod_{a \in \mathcal{A}} p_{s}(a)^{\vec{n}_{i, s}(a)},     
\end{align}
where $\vec{n}_{i, s}$ is the \emph{count vector} associated with suffix $s$
\begin{align}\label{eqn:counting-vector}
    \vec{n}_{i,s}(a) & := \text{ the number of times symbol } a \in \mathcal{A} \notag\\
    &\text{ follows suffix } s \text{ in sequence } (x_1, \ldots, x_i). 
    \vspace{-0.1cm}
\end{align} 

Leveraging the multiplicative nature of the likelihood function, \cite{willems1995context} proposed the context tree weighting (CTW) algorithm. CTW estimates the probability of the sequence $x_{1}^{n}$ by the auxiliary parameters $p^{e}_{n, s}, p^{w}_{n, s}$'s, where $e$ stands for ``estimation", and $w$ stands for ``weighted":  
\begin{enumerate}[topsep=-2pt, itemsep=-2pt]
\item For each string $s$ with $|s| \leq D$, compute 
{\small
\begin{align*}
&p^{e}_{n, s} = \frac{ \Gamma(\sum_{a \in \mathcal{A}} \boldsymbol{\alpha}(a) ) }{ \Gamma(\sum_{a \in \mathcal{A}} ( \vec{n}_{s}(a) + \boldsymbol{\alpha}(a)) }\prod_{q \in \mathcal{A} } \frac{ \Gamma( \vec{n}_s(a) + \boldsymbol{\alpha}(a) ) }{ \Gamma(\boldsymbol{\alpha}(a)) },
\end{align*}}
where $\vec{n}_{s}=\vec{n}_{n,s}$, %is the counting vector $\vec{n}_{i, s}$ with $i = n$,  
$\Gamma(\cdot)$ is the Gamma function, and $\boldsymbol{\alpha}$ is a prior-related vector to be specified shortly. $p^{e}_{n, s}$ is computed from the statistics for that suffix $s$, i.e., $\vec{n}_s$.
\item From nodes in the $D$-th level to the $0$-th level (i.e., root), iteratively compute 
{\small
\begin{align*}
p^{w}_{n,s}:=
\left\{
\begin{array}{ll}
p^{e}_{n,s}, &\mbox{if $|s| = D$,}\\
\lambda p^{e}_{n,s} + (1-\lambda) \prod_{q \in \mathcal{A}} p^{w}_{n,qs}, &\mbox{otherwise,}
\end{array}
\right. 
\end{align*}}
where $qs$ is the string by appending symbol $q \in \mathcal{A}$ before the suffix $s$. This recursion weights $p^e$ at a node with $p^w$'s at its children nodes.
\end{enumerate}
We illustrate this iterative computation and the corresponding information flow in Fig. \ref{fig:main}. 

\cite{kontoyiannis2022bayesian} took the Bayesian view, and showed that the probability $p^{w}_{n,()}$ computed at the root has a Bayesian interpretation. A prior distribution $\pi_{\text{CTW}}$ is introduced, over the CTs in the collection of CTs with a depth at most D, which is denoted as $\mathcal{T}(D)$. 
Recall that a CT consists of the tree structure $T$ and the set of the transition distributions at the leaves $\{p_s \in \Delta_{\mathcal{A}}\}$. The prior distribution is given as $\pi_{\text{CTW}}(T, (p_{s})_{s \in \mathcal{L}(T)} ) = \pi_D(T) \prod_{ s \in \mathcal{L}(T) } \pi_p( p_{s} )$, where $\pi_D(\cdot)$ represents a bounded branching process that each node at a level lower than $D$ stops branching with probability $\lambda$ or branches to $|\mathcal{A}|$ children with probability $(1 - \lambda)$; and $\pi_{p}(p_{s})$ is the Dirichlet distribution: 
\begin{align}
    &\pi_D(T)  = (1 - \lambda)^{( |\mathcal{L}(T)| - 1 ) / (A - 1)} \lambda^{ |\mathcal{L}(T)| - |\mathcal{L}_D(T)| }, \notag\\
    &\pi_{p}(p_s) = \text{Dir}(p_s; \{ \boldsymbol{\alpha}(a) \}_{a \in \mathcal{A}} ),
\end{align}
where $\mathcal{L}_{D}(T)$ is the set of the leaves of $T$ at depth $D$ and $\{ \boldsymbol{\alpha}(a) \}_{a \in \mathcal{A}}$ are the Dirichlet parameters. A typical choice is $\boldsymbol{\alpha}(a) = 0.5$, corresponding to the Jeffreys prior. The Bayesian view implies that the sequences can be viewed as being generated hierarchically, as illustrated in Fig. \ref{fig:data-generation} in the appendix.

\begin{theorem} \citep[Theorem 3.1]{kontoyiannis2022bayesian} 
\label{lem:predicteD-likelihood}
$p^{w}_{n,()}$ computed by CTW equals the Bayesian predicted probability under the prior $\pi_{\text{CTW}}$ specified by $(D, \lambda, \boldsymbol{\alpha})$:
\begin{align*}
    &p^{w}_{n,()}  = P_{\pi_{\text{CTW}}}(x_{1}^{n} | x_{1-D}^{0} )\\
    &=\sum_{T \in \mathcal{T}(D)} \int P_{T, \{p_s\}}(x_{1}^{n} | x_{1-D}^{0}) \pi(T, \{p_s\} ) \prod_{s \in \mathcal{L}(T)} \mathrm{d} p_{s} .
\end{align*}
\end{theorem}

This implies that CTW computes exactly the probability of sequence $x_{1}^{n}$ under the prior $\pi_{\text{CTW}}$ parameterized by $(D, \lambda, \boldsymbol{\alpha})$.
%and , i.e., the $p^{w}$ at the root. 
The next-token probability can the be computed sequentially as
\begin{align}
&P_{\pi_{\text{CTW}}}(x_{i+1} | x_{1-D}^{i}) \notag\\
&\qquad= {P_{\pi_{\text{CTW}}}(x_{1}^{i+1} | x_{1-D}^{0})}/{P_{\pi_{\text{CTW}}}(x_{1}^{i} | x_{1-D}^{0})}\label{eqn:autoregressive}
\end{align}
which achieves a code length at most $\lceil \log_2(1/p^{w}_{n,()}) \rceil$ \citep{willems1995context,willems1997complexity}, i.e., Bayesian optimal. 

\vspace{-0.1cm}
\subsection{Analysis of Attention Maps} \label{sec:analysis-attention-maps}
\vspace{-0.1cm}

\begin{figure*}[htb]
\centering
 \includegraphics[width=0.7\textwidth]{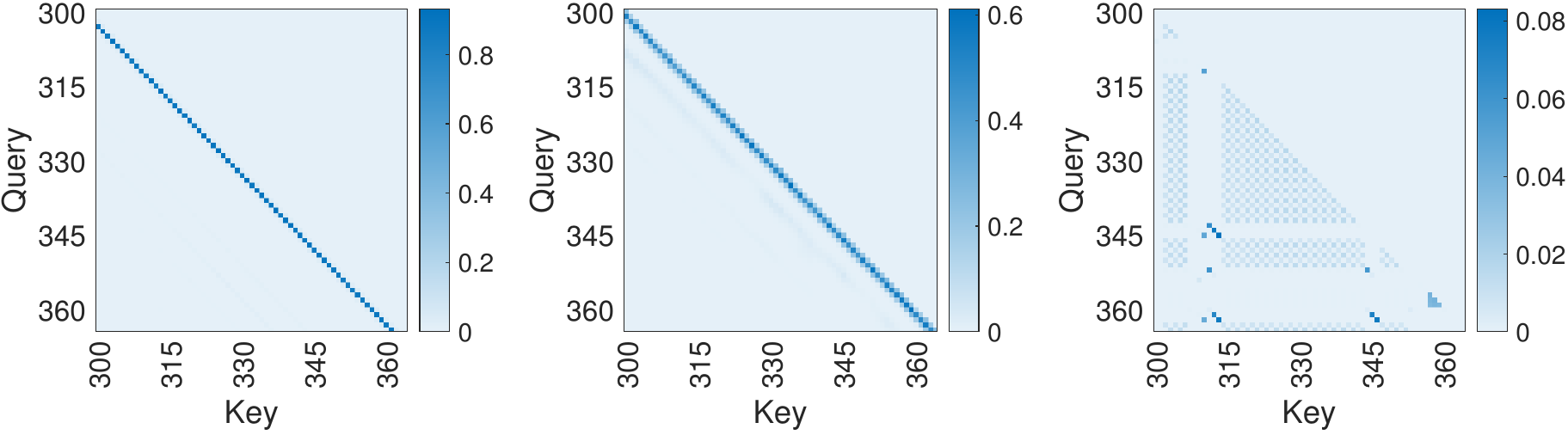} 
  \includegraphics[width=0.82\textwidth]{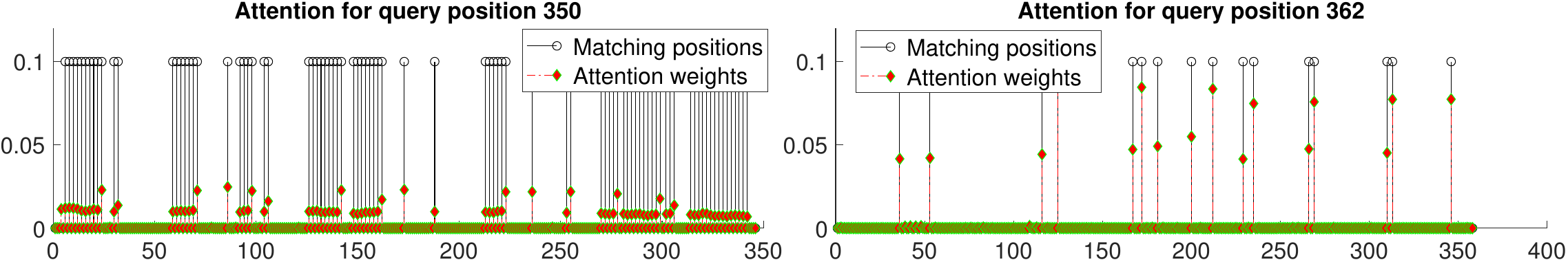}
 \caption{Top: Attention heatmaps. Bottom: Matching between suffix locations and attention values.\label{fig:heatmap}}
 \vspace{-0.3cm}
 \end{figure*}

We analyzed the attention maps of the trained transformers, where a few distinguished patterns emerge. One pattern is solely relative-position dependent. In the first two panels of Fig. \ref{fig:heatmap} (top), we observe off-diagonal stripes for these two attention heads, which are a few positions below the main diagonal. They can be a single off-diagonal or a collection of several off-diagonals. This indicates that the query position is attending positions at a few fixed but close distances ahead of itself. This pattern usually appears in the first or second layers of the transformers. Combining with the suffix structure in compression algorithms such as CTW, such an attention pattern suggests the suffix is being copied into the current query position for subsequent processing. The off-diagonal stripes may have a width greater than $1$, as shown in the second panel, which can be viewed as copying a mixture of the tokens in the suffix, suggesting the flexibility of transformers in forming certain ``soft" suffixes. 

Another pattern, shown in the third panel, has more sophisticated spotty patterns, and the attention appears to depend more explicitly on the current token features instead of the position alone, and they usually appear in the second layer or above in the transformers. Taking query positions 350 and 362 for the attention head shown in the third panel of Fig. \ref{fig:heatmap} (top), we plot in Fig. \ref{fig:heatmap} (bottom) the positions in the data sequence that match their suffixes of length-$3$ using the stem plots with a black circle on top, and the attention values as the red stems with the diamonds on top. The positions match perfectly, though the attention weights have some variations among them. This attention pattern suggests that it is collecting statistical information for those positions with the matched suffix. Several attention heads present similar patterns but with different suffix lengths. 

\vspace{-0.1cm}
\subsection{Interpretation via Constructions} \label{sec:attention-map}
\vspace{-0.1cm}

We next connect transformers' ICL performance to that of CTW via constructions. 

\vspace{-0.1cm}
\paragraph{A New Representation of CTW.} The original CTW algorithm computes the probability of the whole sequence for a particular next token realization. Though there exist studies for sequential computation in the literature \cite{willems1997complexity,willems1997complexity2,willems1998reducing}, only binary sequences were considered and they are incompatible with the transformer architecture. To construct meaningful transformers, we propose a novel representation of the predictive probability $P_{\pi_{\text{CTW}}}(x_{n+1} | x_{1-D}^{n})$ in the theorem below, based on a weighted blending of the next token prediction probability vectors for each potential suffix $s_{n, l} := x_{n-l+1}^n$ of length $l = 0, 1, \ldots, D$. 
The proof is given in Appendix \ref{app:new-formula}.

\begin{theorem}\label{thm:main-new-formula}
The next token probability is given as
\begin{align*}
    &\vec{P}_{\pi_{\text{CTW}}}(x_{n+1} | x_{1}^n) = \sum_{l = 0, \ldots, D}~  \omega_{n, l} \cdot \vec{p}_{n, s_{n, l}}(x_{n+1}), 
\\&\qquad \qquad\text{ where }
\vec{p}_{n, s_{n, l}}(a) := \frac{ \boldsymbol{\alpha}(a) + \vec{n}_{n,s_{n,l}}(a) }{ \sum_{q} (\boldsymbol{\alpha}(q) + \vec{n}_{n,s_{n,l}}(q)) },
\end{align*}
and $\boldsymbol{\omega_{n}} \in \Delta_{D+1}$ which are defined recursively using 
\begin{align*}
&\ln(\omega_{n, l}) - \ln(\omega_{n, l-1})  = \ln(1-\lambda) - \mathbb{I}_{l = D}\ln(\lambda) \\
&\quad+ \ell^{e}_{n, s_{n, l}}  - \ell^{e}_{n, s_{n, l-1}}+ \sum_{q \in \mathcal{A}} \ell^{w}_{n, q s_{n, l-1}} - \ell^{w}_{n, s_{n, l}}, 
\end{align*}
for $l = 1,\ldots, D$, where $\ell^{e}_{n, s} := \ln(p^{e}_{n, s})$, $\ell^{w}_{n, s} := \ln(p^{w}_{n, s})$, and $\mathbb{I}_{(\cdot)}$ is the indicator function. 
\end{theorem}

\begin{figure*}[t!]
\centering
\includegraphics[width=0.8\textwidth]{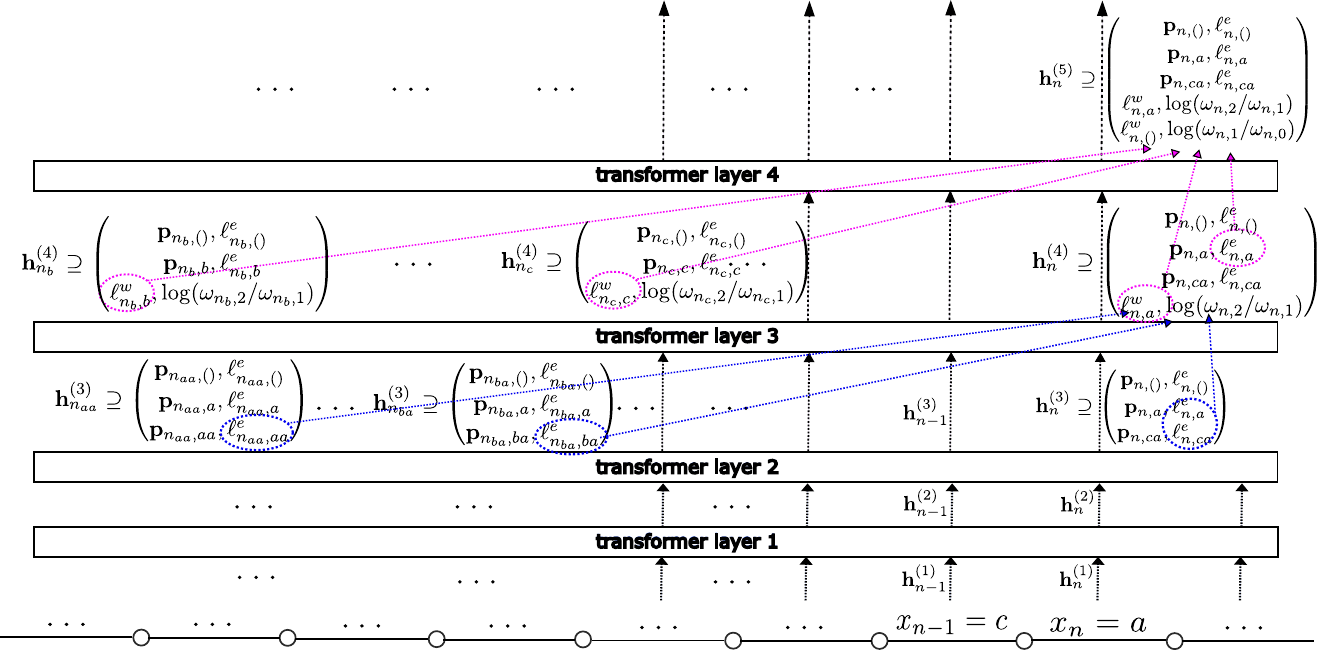}
\caption{The last $D$ layers mimic the CTW algorithms, and $\omega_{n,\cdot}$'s collect information over different locations.\label{fig:higherlayers}}
\vspace{-0.2cm}
\end{figure*}

Consider an example where $x_{n-1}=c$, $x_n=a$, and $D=2$. Each suffix $s_{n, l}$, e.g., $s_{n,0} = (), s_{n,2}=ba$, can potentially be the true suffix for the next token, i.e., $s_{n, l} \in \mathcal{L}(T)$. The Bayesian optimal next token prediction for $s_{n, l}$ is $\vec{p}_{n, s_{n, l}}$, and the weight $\omega_{n, l}$ assigns a ``credibility'' that $s_{n, l}$ is the true suffix. Theorem \ref{thm:main-new-formula} states that these weights use information on the suffix path (e.g., root-$a$-$c$ in this example) and from their siblings $p^{w}_{n, qs_{n, l-1}}$, but it is independent of the exact realization of the next token (see also Fig. \ref{fig:higherlayers}). One obvious advantage of this representation is that, unlike the original CTW algorithm, which performs the computation specifically for the next token realization (\ref{eqn:autoregressive}), the new representation computes the whole probability distribution (vector) and avoids repeated computation of many identical quantities. The computation of $\omega_{n,\cdots}$'s in the transformer architecture is non-trivial, which will be given shortly. 

\vspace{-0.1cm}
\paragraph{Transformer Construction for CTW: }We next provide a construction of $(2 + D)$-layer transformer with sufficient representation power in the FF layer that can essentially approximate CTW, which demonstrates the capacity of transformers. The first two layers are motivated by the attention map patterns observed in Section \ref{sec:analysis-attention-maps}, which collect statistics needed for Theorem \ref{thm:main-new-formula}; the last $D$ layers are induction layers imitating the CTW procedure. 

We assume a one-hot initial embedding, with additional scratchpad elements initialized as zeros and a positional embedding, i.e., $\vec{h}_i^{(1)}=(\vec{x}_i; \vec{0}; \vec{pos}_{i})$ where $\vec{x}_i \in \mathbb{R}^{A}$ denotes the one-hot (column) embedding of $x_i$, $\vec{pos}_{i} = (1, \cos(i \pi / N), \sin(i \pi / N))^\top$ is a sinusoidal positional embedding, and the remaining $(E-A-3)$ elements being zero ($E$ to be specified later).

We begin with the first layer, which is referred to as a finite-memory context-extension layer. 
\begin{theorem}
\label{thm:extension}
There is an $M$-headed transformer layer that performs finite-memory context-extension with the following output, using the initial one-hot embedded input $\vec{H}^{(1)}$:
\begin{align}
\vec{h}^{(2)}_i = (\vec{s}_{i, M+1}; \vec{0}; \vec{pos}_i), 
\label{eq:finite-memory}
\end{align}
where $\vec{s}_{i, M+1} = (\vec{x}_i;\ldots;\vec{x}_{i-M})$ is the vector version of the $M$-length suffix $s_{i, M+1} = x_{i - M}^{i}$. 
\end{theorem}
This layer essentially copies $M$ past embedded symbols to the current position $i$, and stacks them below the current symbol $\vec{x}_i$. This operation utilizes the positional encoding $\vec{pos}_i$ via rotation and matching the corresponding positions. Consistent with our empirical observations, we use multiple heads that scale with the parameter $D$, instead of complex embedding (e.g., embedding history in a high-precision scalar floating number) or that scales as $|\mathcal{A}|^D$. The proof is given in Appendix \ref{app:extension}. 

The second layer is referred to as the statistics collection layer, which takes a sequence of vectors $\vec{h}^{(2)}_i$, $i=1,\ldots,N$, defined in (\ref{eq:finite-memory}) as its input. To rigorously specify the function of this layer, we define the $k$-gram (forward) statistics vector $\vec{g}_{i, s}$ with $|s| = k-1$, which in plain words, is the empirical probability distribution of the next token associated with the suffix $s$ for sequence $x_{1}^{i}$. Similarly, we define the $k$-gram backward statistics vector $\vec{g}_{i-1, s}^{\leftarrow}$, which is the empirical probability distribution of the previous token associated with the suffix $s$ for $x_{1}^{i-1}$. Mathematically, $\forall a \in \mathcal{A}$ %for a suffix $s$ and position $i$, 
\begin{align*}
    &\vec{g}_{i, s}(a) = \frac{\vec{n}_{i, s}(a)}{ \sum_{q \in \mathcal{A}} \vec{n}_{i, s}(q)}, \, 
    \vec{g}_{i-1, s}^{\leftarrow}(a) = \frac{ \sum_{q \in \mathcal{A}} \vec{n}_{i, as}(q) }{ \sum_{q \in \mathcal{A}} \vec{n}_{i, s}(q) },
   % &\qquad \forall a \in \mathcal{A}, %\label{eqn:forward-backward-statistics}
\end{align*}
where $\vec{n}_{i, s}$ is the counting vector defined in (\ref{eqn:counting-vector}), and $\sum_{q \in \mathcal{A}} \vec{n}_{i, s}(q)$ is the number of appears of the string $s$ in the sequence $x_{1}^{i-1}$. For both $\vec{g}_{i, s}$ and $\vec{g}_{i-1, s}^{\leftarrow}$, if the suffix $s$ has not appeared in data $x_{1}^{i-1}$, it can be initialized arbitrarily as a vector in the probability simplex. 

\begin{figure}[thb!]
\centering
\includegraphics[width=0.35\textwidth]{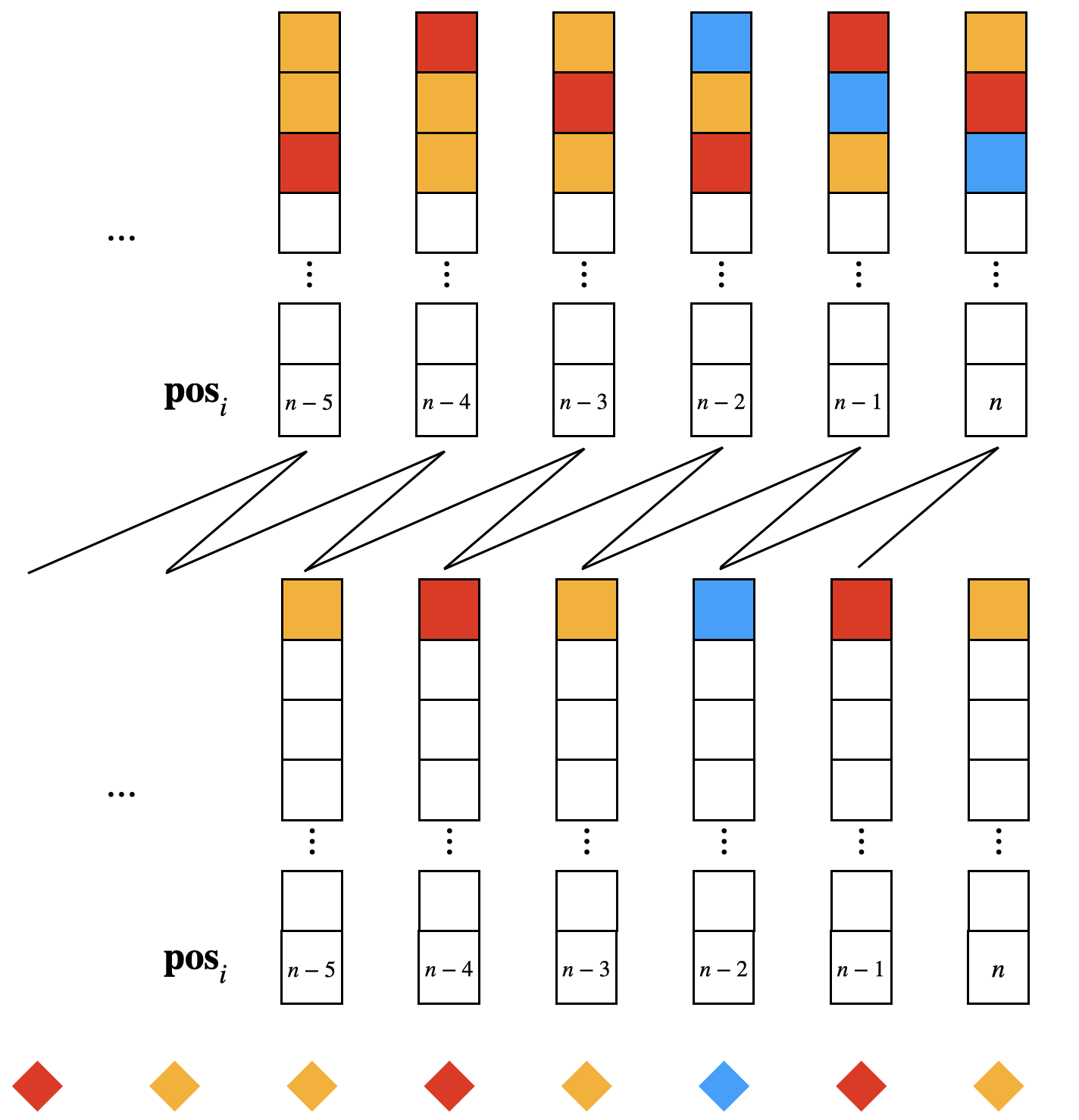}
\includegraphics[width=0.35\textwidth]{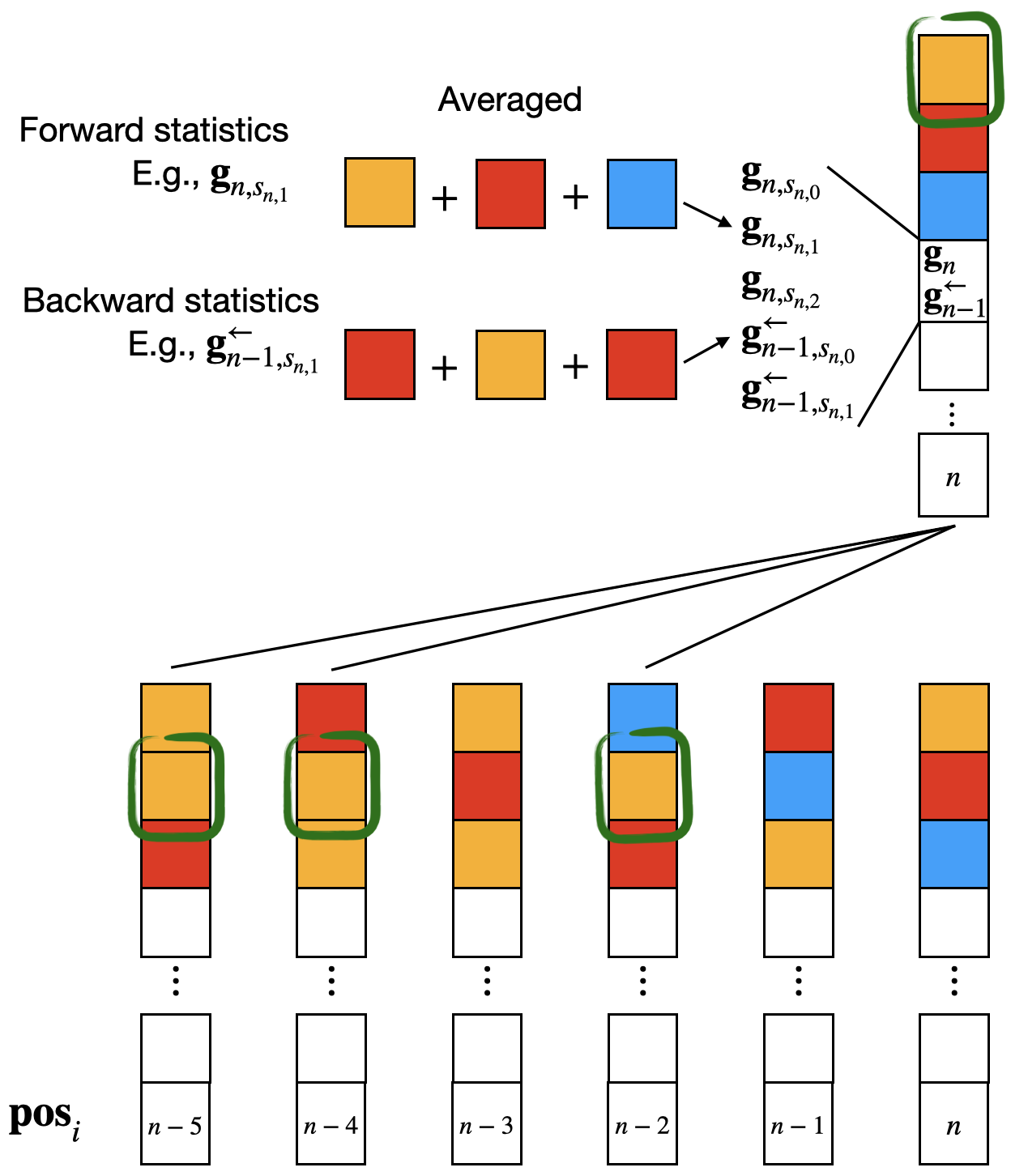}
\caption{The finite-memory context-extension layer and the statistics collection layer for $D=2$.}
\label{fig:construction-first-second}
\vspace{-0.3cm}
\end{figure}

We have omitted the dimensionality of several zero matrices when they are obvious from the context. The first two layers are illustrated in Fig. \ref{fig:construction-first-second}.

\begin{theorem}
\label{thm:statistics}
There is an $M'$-head attention layer, where $M'\leq M+1$, which can collect statistics, defined by the following output, with $\vec{H}^{(2)}$ in (\ref{eq:finite-memory}) as its input:
\begin{align}
    \vec{a}^{(2)}_i=(\vec{s}_{i, M+1}; \vec{g}_{i, M'}; \vec{g}^{\leftarrow}_{i-1, M'} ; \vec{0}; \vec{pos}_i), \label{eqn:statistics}
\end{align}
where $\vec{g}_{i, M'} := (\vec{g}_{i, s_{i, 0}}; \ldots;\vec{g}_{i, s_{i, M'-1}})$ and $\vec{g}^{\leftarrow}_{i-1, M'} = (\vec{g}^{\leftarrow}_{i-1, s_{i, 0}};\ldots;\vec{g}^{\leftarrow}_{i-1, s_{i, M'-1}})$.
\end{theorem}
This functional layer essentially collects $k$-gram statistics for various lengths of $k=1,2,\ldots,M'$. For example, when $k=3$, it collects the normalized frequency associated with the suffix $(x_{n-1},x_{n})$. 

\begin{figure}[thb!]
\centering
\includegraphics[clip, trim=0.5cm 0.5cm 0.8cm 0.5cm, width=0.45\textwidth]{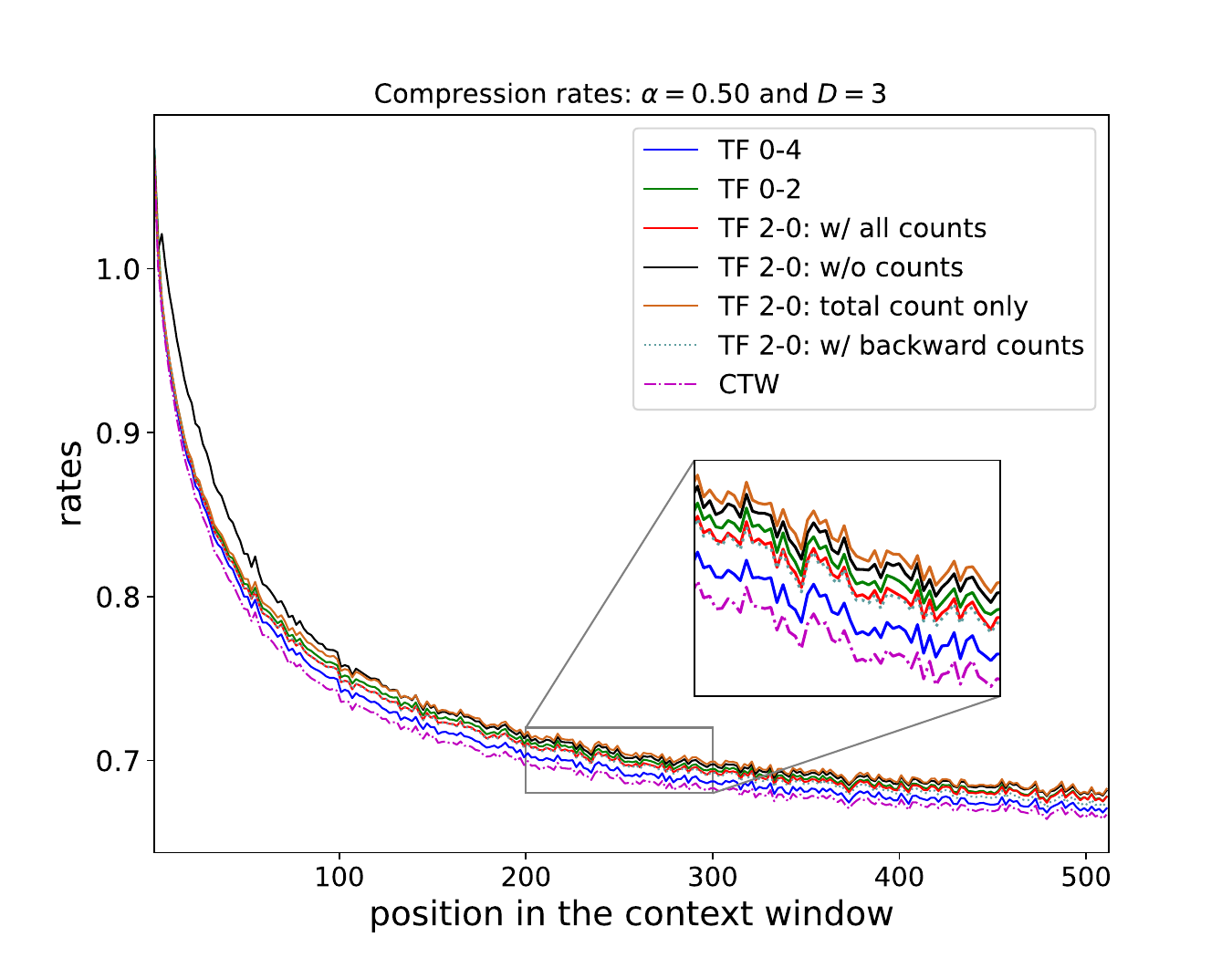}
\caption{The performance of normal and synthetic 2-layer transformers. \label{fig:normalsyn}}
\vspace{-0.3cm}
\end{figure}

It is important to note that for ICL of FOMCs, two layers that collect forward statistics $\vec{g}_{i, M'}$ with $M'=D+1$ are sufficient \citep{edelman2024evolution}. However, for the ICL-VOMC task, the underlying CT structure is unknown; therefore, collecting such simple statistics is \textit{no longer sufficient}, and it is also the reason that the FF layer is important for VOMCs while less so for FOMCs. As indicated in Theorem \ref{thm:main-new-formula}, the information of counting statistics $\vec{n}_{i, s_{i, l}}$ is important since the weights heavily depend on $\vec{n}_{i, s}(a)$. Yet due to the softmax function, only (normalized) probability values can be obtained instead of the counts. With the backward statistics $\vec{g}^{\leftarrow}_{i, s}$, $\vec{n}_{i, s_{i, l}}$ can be derived as 
\begin{align*}
\vec{n}_{i, s_{i, l}}(a) &= \frac{\vec{n}_{i, s_{i, l}}(a)}{ \sum_{q \in \mathcal{A}} \vec{n}_{i, s_{i, l}}(q) } \cdots \\
&\qquad\cdots\frac{ \sum_{q \in \mathcal{A}} \vec{n}_{i, s_{i, 1}}(q)}{ \sum_{q \in \mathcal{A}} \vec{n}_{i, s_{i, 0}}(q) } \sum_{q \in \mathcal{A}} \vec{n}_{i, s_{i, 0}}(q) \\
&= i\cdot\vec{g}_{i, s_{i, l}}(a) \prod_{j = 0}^{l - 1}  \vec{g}_{i - 1, s_{i, j }}^{\leftarrow}(x_{i - j } ). 
\end{align*}
Taking $M = M'-1=D$, after the statistics collection layer, a sufficiently wide FF layer with ReLU activation can compute (by the universal approximation capability) \citep{cybenko1989approximation,hornik1989multilayer}.
\begin{align*}
    \vec{h}^{(3)}_i=(\vec{s}_{i, D+1}; \vec{p}_{i, D} ; \vec{l}^{e}_{i, D}; \ell^{w}_{i, s_{i, D}} ; \vec{0}; \vec{pos}_i),
    %\label{eqn:input-induction}
\end{align*}
where $\vec{p}_{i, D} = (\vec{p}_{i, s_{i, 0}}; \vec{p}_{i, s_{i, 1}}; \ldots; \vec{p}_{i, s_{i, D}})$ and $\vec{l}^{e}_{i, D} = (\ell^{e}_{i, s_{i, 0}}; \ldots; \ell^{e}_{i, s_{i, D}} )$ in Theorem \ref{thm:main-new-formula}.

To fulfill the Bayesian optimal prediction, we use the following induction layer that iteratively computes $\ell^{w}_{i, s}$ for suffixes on the suffix path and their siblings, as well as  the weight differences, which are denoted by $\delta_{i, l} := \ln(\omega_{i, l}) - \ln(\omega_{i, l-1})$ for $l = D, D-1, \ldots, 1$. Specifically, the desired embedding
\begin{align}
    &\vec{h}^{(\ell)}_i=(\vec{s}_{i, M^{(1)}+1}; \vec{p}_{i, D} ; \vec{l}^{e}_{i, D}; \delta_{i, D}; \delta_{i, D-1}; \ldots ; \notag\\
    &\qquad\quad\delta_{i, D - \ell + 4} ; \ell^{w}_{i, s_{i, D+3-\ell}} ; \vec{0}; \vec{pos}_i),% ~ \ell=3,\ldots, 3 + D. 
    \label{eqn:h-ell-induction}
\end{align}
for $\ell = 3, 4, \ldots, 3 + D$. 
\begin{theorem}
\label{thm:induction}
{\hspace{3mm}} There exists an $A$-head transformer \\layer that can perform the induction: Takes $\vec{H}^{(\ell)}$ in (\ref{eqn:h-ell-induction}) as input and outputs $\vec{H}^{(\ell+1)}$, and the final output layer taking $\vec{H}^{(D+3)}$ as input can output the $A$-dim Bayesian optimal vector
$ \sum_{l = 0, \ldots, D}~  \omega_{n, l} \vec{p}_{n, s_{n, l}}.$
\end{theorem}

The last $D$-layer construction is shown in Fig. \ref{fig:higherlayers} with $D=2$, where a recursive structure emerges to collect information from different locations of the context window to compute the weights $\omega_{n,\cdots}$'s; the resemblance to the recursion in Fig. \ref{fig:main} is clear.

\vspace{-0.1cm}
\paragraph{Parallelization and Layers:} A key consideration is for transformers to parallelize the computation at all token positions simultaneously. While a naive construction to perform CTW could certainly process the tokens sequentially as the CTW algorithm suggests, it would require the number of layers to grow linearly with the context window size. Theorem \ref{thm:main-new-formula} is the theoretical tool allowing us to instead perform the same computation in $D+2$ steps, i.e., the $D+2$ layers. 

The construction is certainly not unique. Consider a 2-layer transformer with a large embedding space, where the number of attention heads scales like $|\mathcal{A}|^D$. Each attention head could be precoded to collect the statistics for each possible suffix of length $\leq D$, and all such information can be stored in $\vec{h}_i^{(3)}$, assuming a sufficiently large embedding space (or a more delicate embedding structure in a smaller space). Each $\vec{h}_i^{(3)}$ would then have all the information on the context tree, on which a wide FF layer could perform the full CTW algorithm to produce the next token distribution. We conscientiously forgo constructions like this, because in large models with $\mathcal{A}$ in the tens of thousands and $D$ in the tens or hundreds, scaling as $|\mathcal{A}|^D$ is unrealistic, and the computation of the FF layer would be extremely complex and hard to interpret. More discussions on potential tradeoffs between layers and attention heads can be found in \cite{rajaraman2024transformers}.

\vspace{-0.1cm}
\subsection{Interpreting Two-Layer Transformers} 
\label{sec:why2layer}
\vspace{-0.1cm}

We next consider how two-layer transformers can learn VOMCs in context. Our construction given above requires $D+2$ layers, and we conjecture that the statistical information collected in the second layer can be used to approximate the optimal weights $\omega_{n, l}$ with an FF network. %, because intuitively, these counts are already very informative. 

To verify our conjecture, we implement the first two constructed transformer layers but control the statistics passed onto the FF layer. In Fig. \ref{fig:normalsyn}, ``TF 0-2'' denotes the canonical 2-layer transformer; ``TF 2-0'' is the transformer with two constructed layers with output $\vec{a}_i^{(2)}$ in \eqref{eqn:statistics}, followed by a trainable FF layer (the FF layer in the second layer of the transformer) and an output layer. The version ``TF 2-0 w/o counts'' does not contain $\vec{g}_{i-1, M'}^{\leftarrow}$ or $\vec{pos}_{i}$ in $\vec{a}_i^{(2)}$; the version ``TF 2-0 total counts only'' does not contain $\vec{g}_{i-1, M'}^{\leftarrow}$ in $\vec{a}_i^{(2)}$ and $\vec{pos}_i$ is replaced by the total count $i$; ``TF 2-0 w/ all counts'' replaces $\vec{g}_{i-1, M'}^{\leftarrow}$ and $\vec{pos}_{i}$ with $\{ \vec{n}_{n, s_{n, l}} \}_{l = 0}^{D}$ and $i$. Even though their performances are rather clustered, we can make the following observations: 1) The performances degrade as more counting information is removed from the representation, and the counting information is clearly important, 2) The performances of ``TF 2-0'' and ``TF 2-0 w/ all counts'' almost match exactly, indicating the main function of the backward statistics $\vec{g}^{\leftarrow}_{i-1,M'}$ is to extract the counts, and 3) The performance of the canonical 2-layer transformer is similar to that of the constructed ``TF 2-0'' and ``TF 2-0: w/ all counts'' than those with less counting information. Therefore, the performances of these different implementations confirm the conjecture. Interestingly, \cite{akyurek2024context} proposed a similar learned-weight $2$\&$3$-gram algorithm to interpret how transformers learn finite automata, and suggested it may be optimal. In contrast, we obtain the (suboptimal) architecture by approximating the provable optimal transformer architecture. 

\vspace{-0.1cm}
\section{CONCLUSION}
\vspace{-0.1cm}

We considered transformers' in-context learning of VOMCs. By drawing an analogy of ICL and Bayesian universal compression, we leverage the CTW and KN-smoothing as baselines.  We observed that two-layer transforms can learn VOMCs in context. To understand the mechanism of transformers' ICL ability, we analyzed the attention maps and provided new transformer constructions. A future work is to extend the approach \citep{edelman2024evolution,nichani2024transformers} to study the pretraining dynamics, to understand whether gradient-based training would yield the given constructions. Since attention-only networks do not perform well and multilayer networks are needed in this setting, such an analysis is considerably more difficult and may require a less conventional approach.

\section*{Acknowledgments}

The work of R.D. Zhou and S. Diggavi was supported in part by the DEVCOM Army Research Laboratory under award \# W911NF1720196, and by the National Science Foundation under award DMS-2502536. The work of C. Tian was partly supported by the National Science Foundation via grants DMS-2312173 and ECCS-2433631. The views and conclusions contained in this document are those of the authors and should not be interpreted as representing the official policies, either expressed or implied, of the funding agencies.

%\bibliography{compression.bib}

\clearpage
\appendix

\onecolumn
\aistatstitle{An Information-Theoretic Approach to Understanding Transformers' In-Context Learning of Variable-Order Markov Chains \\Supplementary Materials}

\section{Related Work}\label{app:related-works}

There have been many efforts in studying the ICL capabilities of transformers. A significant recent development is the elucidation of the connection to gradient descent, particularly for linear regression tasks \citep{von2023transformers,akyurek2022learning,dai2023can,ahn2024transformers}. \cite{li2023transformers} formulated the ICL problem as a multi-task learning problem and considered ICL for several simple problem settings for which the authors provide risk bounds for ICL of supervised learning algorithms in these problem settings. \cite{kirsch2022general} viewed the ICL problem as a meta-learner and studied the relation between tasks and model sizes. These approaches focused on the ICL of supervised learning tasks, such as classification and regression, while this work belongs to another direction of studying ICL for the next token prediction of some unknown underlying dynamics. 

\cite{olsson2022context} studied the induction head, i.e., the forming of small $k$-gram attention in LLMs. \cite{reddy2023mechanistic} studied the balance between ICL and in-weights learning, and observed the abrupt emergence of the induction head corresponds to the emergence of ICL. The induction head was generalized to the statistical induction head in \citep{edelman2024evolution} mainly to study bigrams. We adopted it but further allowed more statistical induction heads for more suffixes to be included together, in the first two layers of the idealized transformer.

There have also been efforts to study transformers and the learning of Markov chains. \cite{xie2021explanation} viewed ICL as a Bayesian inference problem, where a latent concept determines an HHM, and the observations from the HHM can lead to the identification of the hidden concept. They studied the eventual ICL capability, i.e., when the number of in-context examples goes to infinity. \cite{hu2024limitation} studied the limitations of transformers on learning to perform belief inference for HMM sources compared to recurrent neural networks. The work in \citep{bietti2024birth} allowed a fixed-order Markov chain to switch to a new deterministic mode, and the authors studied the training behavior of the corresponding ICL task with this mode transition. \cite{akyurek2024context} made a comprehensive empirical comparison of various language models on random finite automata, and showed that the transformer performs the best among these models. \cite{makkuva2024attention} studied the loss landscape during transformer training on sequences generated from a single fixed-order Markov chain, using a single-layer transformer. Neither of these studies considered ICL. 

More recently, \cite{rajaraman2024transformers} considered ICL of FOMCs with single-head transformers, and provided a construction to show that it is possible to use a single attention head to capture longer memory in the sequence. An empirical study on transformers' ICL capability for $n$-grams was conducted in \cite{svete2024can}. Learning certain special Markov sources was studied from the perspective of tokenization usage in \cite{rajaraman2024analysis}, which showed that tokenizations can play an important role in the overall architecture. Particularly, it was observed that transformers may become difficult to train without a tokenizer, while training became easier with a tokenizer. The work most relevant to us is \citep{edelman2024evolution}, where ICL of a fixed-order Markov chain was considered, and the training behavior was studied both empirically and theoretically, and the formation of induction heads in a two-layer network was demonstrated. Another closely related work is \cite{nichani2024transformers}, where a more general dependence graph structure among positions in the context window is allowed. 

All these existing work assumed fixed-order Markov models or fixed-order HHMs, usually with orders kept at 1 or 2; moreover, they almost all focus on the emergence of the induction heads during training or the training landscape. Our study is different firstly in the variable-order nature of the Markov models, and secondly the focus on the on-time ICL performance instead of the training landscape and behavior. 

Lossless data compression has a long history, with many different algorithms being developed over the years. The most popular general-purpose compression algorithms are perhaps the Lempel-Ziv compression algorithms \citep{ziv1977universal,ziv1978compression} and their variants, which belong to dictionary-based compression algorithms. These algorithms do not explicitly maintain any probabilistic models, and their efficiency comes from maintaining an efficiency dictionary of sequences that have been seen before, and are matched with future sequences. More powerful compression algorithms usually maintain probability models explicitly, which are then plugged into an arithmatic coding module \citep{Rissannen76,Pasco-76,rissanen1979arithmetic} for efficient compression. The most well-known classes of algorithms in this category are the context-tree weighting algorithm \citep{willems1995context,begleiter2004prediction,kontoyiannis2023context} and prediction by partial matching \citep{cleary1984data}. The former enjoys a strong theoretical guarantee, particularly on binary sources \citep{willems1995context}, but has some difficulty in its practical implementation \citep{willems1998context,willems1996context,sadakane2000implementing,begleiter2004prediction}, particularly for large alphabet sizes and sequential data. The latter is based more on heuristics, and has been improved and extended in various ways \citep{cleary1997unbounded,moffat1990implementing,shkarin2002ppm}. Methods based on probabilistic modeling are usually more resource-intensive, though they have gained more popularity recently due to the increased availability of computing resources. The evaluation given in \citep{begleiter2004prediction} suggests that CTW and PPM are the two most powerful compression algorithms in practice. There are other compression algorithms, such as those based on the Burrows-Wheeler transformation \citep{burrows1994block}, which does not explicitly maintain a probabilistic model, but are also not dictionary-based. 

Universal compression is often used jointly with arithmetic coding (AC) \citep{rissanen1979arithmetic,Pasco-76}, which requires an estimated probability distribution $\hat{P}$ for the next symbol to perform compression. AC is complicated, nevertheless, here it suffices to view it as a blackbox that compresses a symbol $x$ with approximately $\ln(1/\hat{P}(x))$ nats, resulting in a rate roughly equal to the cross-entropy between $\hat{P}$ and $P$, that latter of which is the true distribution. 

\section{The PPM Algorithm} \label{app:ppm}
The PPM algorithm (with finite memory of parameter $D_{\text{PPM}}$) blends several CTs by utilizing an escape symbol (Esc), and adaptively refines the CT model using the observed samples. The key idea is that the estimated probability distribution for an emitted symbol is only used when there were past observations of this string. For other cases, the escape symbol is encoded, indicating a shorter suffix needs to be used.

\begin{table*}[th!]
\vspace{-0.1cm}
\setlength\tabcolsep{3.3pt}
\centering
\caption{PPM counts after observing string $(a,b,c,a,b,b,c)$\label{tab:PPM}}
\resizebox{0.65\textwidth}{!}{
\begin{tabular*}{0.8\textwidth}{ccccc|ccccc|ccccc|ccccc}%{*{20}{c}}
\toprule
\multicolumn{5}{c|}{order $k=2$} & \multicolumn{5}{c|}{order $k=1$} & \multicolumn{5}{c|}{order $k=0$} & \multicolumn{5}{c}{order $k=-1$} \\ 
\multicolumn{3}{c}{prediction}& $c$ & $p$ &\multicolumn{3}{c}{prediction} & $c$ & $p$ & \multicolumn{3}{c}{prediction} &$c$ & $p$ &\multicolumn{3}{c}{prediction} &$c$ & $p$\\ \hline\hline
$(a,b)$ & $\rightarrow$ & $b$ & $1$ & $\frac{1}{3}$ & $a$  & $\rightarrow$ & $b$ & $2$ & $\frac{2}{3}$ &   & $\rightarrow$  & $a$ & $2$ & $\frac{1}{4}$ &  &  &  &   & $\frac{1}{|\mathcal{A}|}$ \\ 
& $\rightarrow$ & $c$ & $1$ & $\frac{1}{3}$       &     & $\rightarrow$ & $\text{Esc}$ &$1$ & $\frac{1}{3}$ &  & $\rightarrow$  & $b$ & $3$ & $\frac{3}{8}$ &  &  &  & \\ 
& $\rightarrow$ & $\text{Esc}$ & $1$ & $\frac{1}{3}$      &$b$  & $\rightarrow$ & $b$  &$1$ & $\frac{1}{4}$  &  & $\rightarrow$  & $c$ & $2$ & $\frac{1}{4}$  &  &  &  & \\ 
$(b,b)$& $\rightarrow$ & $c$ & $1$ & $\frac{1}{2}$&     & $\rightarrow$ & $c$  &$2$ & $\frac{1}{2}$  &  & $\rightarrow$ & $\text{Esc}$ & $1$ & $\frac{1}{8}$  &  &  &  & \\ 
& $\rightarrow$ & $\text{Esc}$ & $1$ & $\frac{1}{2}$      &     & $\rightarrow$ & $\text{Esc}$ &$1$ & $\frac{1}{4}$  &  &                 &    &     &                 &  &  &  & \\ 
$(b,c)$& $\rightarrow$ & $a$ & $1$ & $\frac{1}{2}$&  $c$& $\rightarrow$ & $a$  &$1$ & $\frac{1}{2}$  &  &                 &    &     &                 &  &  &  & \\ 
       & $\rightarrow$ & $\text{Esc}$& $1$ & $\frac{1}{2}$&     & $\rightarrow$ & $\text{Esc}$ &$1$ & $\frac{1}{2}$  &  &                 &    &     &                 &  &  &  & \\
$(c,a)$& $\rightarrow$ & $c$ & $1$ & $\frac{1}{2}$&     &               &      &    &                &  &                 &    &     &                 &  &  &  & \\ 
       & $\rightarrow$ & $\text{Esc}$& $1$ & $\frac{1}{2}$&     &               &      &    &                &  &                 &    &     &                 &  &  &  & \\\hline
\end{tabular*}
}

\end{table*}
%\end{wraptable}

We illustrate this context tree blending approach by the example shown in Table. \ref{tab:PPM}, where $\mathcal{A}=\{a,b,c\}$, and the memory length $D_{\text{PPM}}=2$. The escape pattern is assigned a count of one (method-A in \citep{moffat1990implementing}). Suppose the next symbol to emit is $a$, then the probability prediction is $\frac{1}{2}$ from the $k=2$ column; if on the other hand, the next symbol to emit is $b$, then the escape symbol is first encoded with probability $\frac{1}{2}$ since there is no string of $(b,c,b)$ in the history, and then we check the column $k=1$, and see that another escape symbol will be encoded since there is also no $(c,b)$, and finally $b$ will be encoded at $k=0$, and the eventual effective probability for $b$ is $\frac{1}{2}\cdot\frac{1}{2}\cdot\frac{3}{8}$. 

Various refinements of the probability estimation can be adopted to further improve the performance, e.g., the exclusion rule, and other methods to initialize the probability for Esc; see e.g. \citep{moffat1990implementing,cleary1997unbounded,begleiter2004prediction}. As the number of observed samples accumulated, all patterns of $(D_{\text{PPM}}+1)$-grams will be observed at least once and the probability prediction will solely based on the column of maximum order $k = D_{\text{PPM}}$. 

\section{Kneser–Ney Smoothing}
\label{app:kn}
Kneser–Ney smoothing is a method primarily used to calculate the next token probability distribution of $(D_{\text{kn}}+1)$-grams. It uses absolute discounting by subtracting a fixed value from the probability's lower-order terms to omit n-grams with lower frequencies. 

First, consider $D_{\text{kn}}=1$, i.e., Kneser–Ney smoothing based on bigram. Let $c(w,w')$ be the number of occurrences of the token $w$ followed by the token $w'$ in the context window before the current position, for any $w,w'\in \mathcal{A}$. Then the next token prediction using the bigram is given by
\begin{align}
p_{\text{kn}}(x_{i+1}|x_i)=\frac{\max(c(x_{i},x_{i+1})-\delta,0)}{\sum_{w'\in\mathcal{A}}c({x_{i},w')}}+\frac{\delta|\{w':c(x_{i},w')>0\}|}{\sum_{w'\in\mathcal{A}}c({x_{i},w')}}p_{\text{kn}}(x_i),
\end{align}
where $P_{\text{kn}(w_i)}$ can be the standard unigram probability distribution estimate until current position or any other estimate of the unigram distribution of the same manner, $\delta\in (0,1)$ is a hyperparameter that can be optimized. For KN-smoothing of order-$D$, the estimate can be computed recursively as
%\newpage
\begin{align}
p_{\text{kn}}(x_{i+1}|x^{i}_{i-D+1})=&\frac{\max(c(x^{i}_{i-D+1},x_{i+1})-\delta,0)}{\sum_{w'\in\mathcal{A}}c(x^{i}_{i-D+1},w')}\notag\\
&\qquad\qquad\qquad+\frac{\delta|\{w':c(x^{i}_{i-D+1},w')>0\}|}{\sum_{w'\in\mathcal{A}}c(x^{i}_{i-D+1},w')}p_{\text{kn}}(x_{i+1}|x^{i}_{i-D+2}),
\end{align}
where $c(\vec{w},w')$ is the occurrences of $(D+1)$-gram $(\vec{w},w')$ until the current position. The hyperparameter $\delta$ controls how the estimates of the different models are blended, with a large $\delta$ being a higher preference for the lower-order estimate. The estimate almost reduces to a fixed-order estimate when there are more than a few previous observations of the same $(D_{\text{kn}}+1)$-grams.

\section{Training Details} \label{app:pretrain-details}

\begin{figure}
  \begin{center}
  \includegraphics[width=0.5\textwidth]{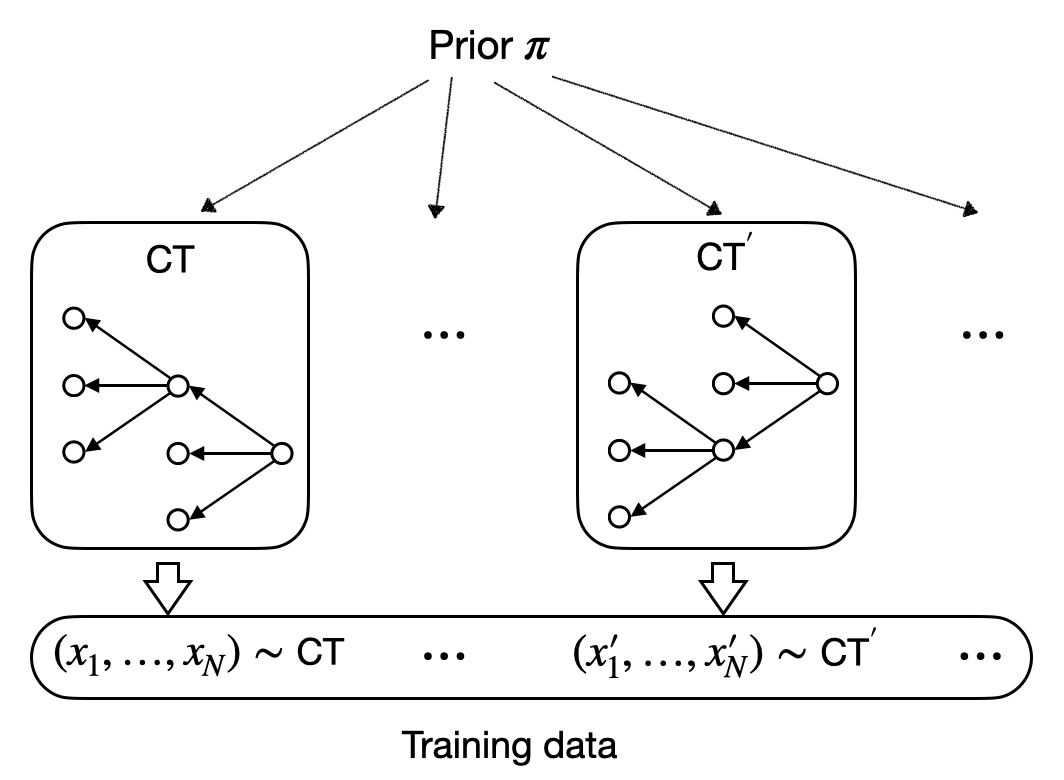}
  \end{center}
  \caption{\small Training data generation \label{fig:data-generation}}
\end{figure}

We choose the alphabet size to be $|\mathcal{A}|=3$ in the experiments. For training, we randomly generate $K=20000$ CTs of various depths (maximum order $D\leq 5$), and then for each CT leaf, we generate a probability distribution. Two different ways of generating these probability distributions are taken: the first approach is to use the Dirichlet distribution to sample such distributions, and the second approach is to randomly select some of the elements in the alphabet to have probability zero, and the others with i.i.d. random values before normalization. Different values of the Dirichlet parameter are tested but only the results do not appear to be sensitive to the choice. For each CT, a source sequence of a certain length (e.g., $N_k=5120$) is produced. The context window $N$ can vary, but in most cases, we set it at $512$ (except when $D=5$, we set it to be $1536$ to allow sufficient data collection in context). Each source sequence is segmented into $\lfloor N_k/N\rfloor$ training sequences. 

During testing, we randomly generate multiple ($8192$ in our experiments) new CTs of varying depths using the same procedure, and for each CT, a sequence of length $N_k=5120$ is generated, and then again segmented into a length of the context window for testing. 

The transformer model is implemented using Pytorch, and trained using the AdamW optimizer with the default parameters. A100/T100 GPUs are used for training. Training a model requires roughly 4 to 6 hours. The batch size is set at 512, and the maximum epoch is set at 100 with early termination allowed after 15 epochs of no improvement. Testing was performed on a local workstation with a GeForce GTX 1660 Ti GPU card.

\section{Additional Experiments on PPM Algorithm}\
\label{app:additional-expriments}

\begin{figure}[h!]
\centering
\includegraphics[width=0.65\textwidth]{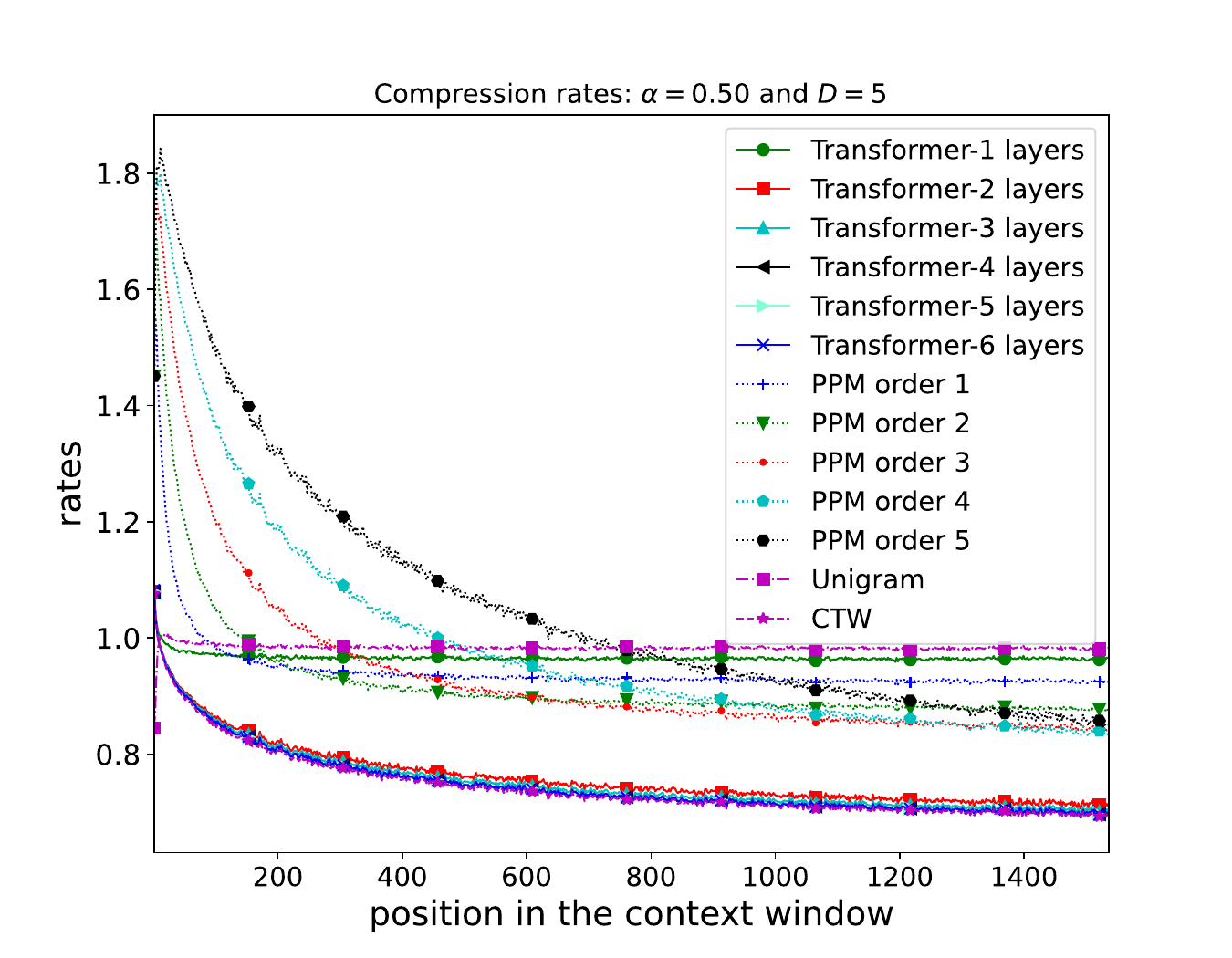}
\caption{Performances of the PPM algorithm in comparison to trained transformers.\label{fig:ppmperformance}}
\end{figure}

We replace the KN-smoothing used in Fig. \ref{fig:tf5} by the PPM algorithm (method A). Since there are many PPM variants, we do not choose and optimize the algorithm in our study, unlike the case for the KN-smoothing, where we optimize the hyperparameter. The results are shown in Fig. \ref{fig:ppmperformance}. It can be seen that the fixed-order estimate behavior is similar between PPM and KN-smoothing, though KN-smoothing appears to be overall more efficient than the specific PPM. For algorithms based on FOMC models, the mismatch between the underlying source and the model is most clear in the latter part of the context window, where the prior and update variations have mostly subsided. In the early part of the context windows, the prior mismatch and sub-optimal probability estimate update both manifest, and it may not be as informative. The similarity between the PPM behavior and the KN-smoothing behavior highlights the fact that the performance gap between CTW (and the transformers) and KN-smoothing (and PPM algorithms) is due to the underlying variable-order and fixed-order modelings, instead of the specific algorithms chosen.

\section{Transformer Architecture} 
\label{app:transformer}

\begin{figure*}[ht]
\centering
\includegraphics[width=0.95\textwidth]{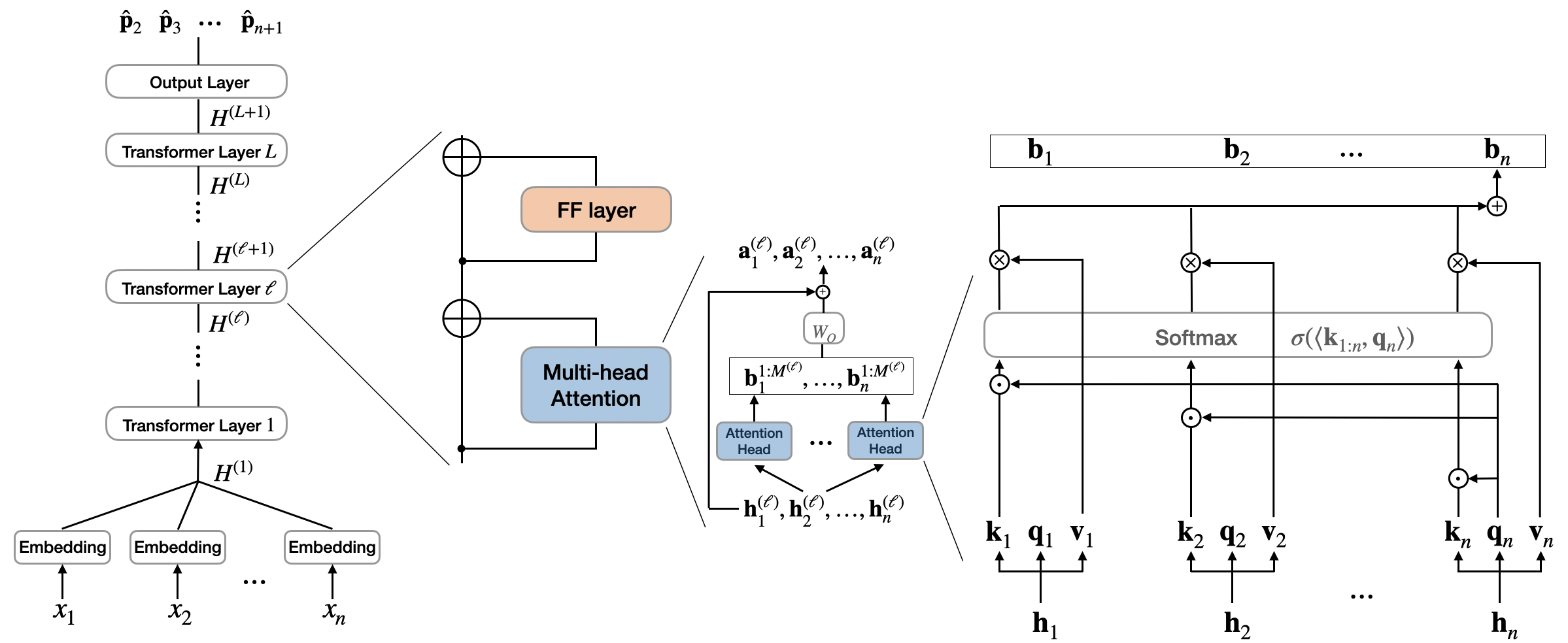}
\caption{Transformer model}
\label{fig:transformer}
\end{figure*}

The transformer architecture used in our construction is illustrated in Fig. \ref{fig:transformer}. 

\section{Proofs of the Theorems for VOMC Sources}

\subsection{A New Formula for Bayesian Next Token Prediction} \label{app:new-formula}

We aim to predict the next token $x_{n+1}$ based on the observations $x_{1-D}^n = (x_{1-D}, \ldots, x_{n})$ via a transformer-friendly formula.  Note that $x_{1-D}^0$ is a place holder or dummy initialization sequence, which does not contain any information of the CT $(T, \{p_s\})$; or alternatively, we can view $P(T, \{p_s\} | x_{1-D}^{0}) = \pi_{\text{CTW}}(T, \{p_s\})$ parameterized by $\lambda, \boldsymbol{\alpha}$. 

\begin{theorem}[Restate Theorem \ref{thm:main-new-formula}] \label{thm:new-formula}
The predicted probability can be computed as
\begin{align}
    \vec{P}_{\pi_{\text{CTW}}}(x_{n+1} | x_{1-D}^n) = \sum_{l = 0, \ldots, D}~  \omega_{n, l} \cdot \vec{p}_{n, s_{n, l}}(x_{n+1}), \label{eqn:next-token}
\end{align}
\vspace{-0.35cm}

where $\vec{p}_{n, s_{n, l}}(a) = \frac{ \boldsymbol{\alpha}(a) + \vec{n}_{n,s_{n,l}}(a) }{ \sum_{q} (\boldsymbol{\alpha}(q) + \vec{n}_{n,s_{n,l}}(q)) }$; and $\omega_{n, \cdot} \in \Delta_{D+1}$ with $\ln(\omega_{n, l}) - \ln(\omega_{n, l-1}) = \ln(1-\lambda) - \mathbb{I}_{l = D}\ln(\lambda) + \ell^{e}_{n, s_{n, l}} - \ell^{e}_{n, s_{n, l-1}} + \sum_{q \in \mathcal{A}} \ell^{w}_{n, q s_{n, l-1}} - \ell^{w}_{n, s_{n, l}}$, where $\ell^{e}_{n, s} = \ln(p^{e}_{n, s})$, $\ell^{w}_{n, s} = \ln(p^{w}_{n, s})$.
\end{theorem}

\textbf{Discussion.} 
Note that $p^{e}_{n, s}, p^{w}_{n, s}$ can be efficiently calculated by the CTW procedure, and compared to calculate $\frac{P_{\pi_{\text{CTW}}}(x_{1}^{n+1} | x_{1-D}^{0})}{P_{\pi_{\text{CTW}}}(x_{1}^{n} | x_{1-D}^{0})}$ for each $x_{n+1}$ the extra computation besides the CTW procedure is $A$ times larger than that by Eq \eqref{eqn:next-token}. The weighted average formula in Eq \eqref{eqn:next-token} gives a natural interpretation for the Bayesian optimal next token predicted probability. Each suffix along the root the leaf path  $s_{n, 0} - s_{n, 1} - \cdots - s_{n, D}$ can potentially be the true suffix, i.e., $s_{n, l} \in \mathcal{L}(T)$, and $\vec{p}_{n, s_{n, l}}$ is in fact the Bayesian optimal next token prediction given $s_{n, l} \in \mathcal{L}(T)$. 

The blending weights $\omega_{n, l}$'s are based on stopping probability $\lambda$, the information in the potential suffix path such as $p^e_{s_{n, s_{n, l}}}$ as well as the information from their siblings $p^{w}_{n, qs_{n, l-1}}$. We can interpret $p^{e}_{n, s}$ as the evidence (unnormalized likelihood) that $s \in \mathcal{L}(T)$, and $p^{w}_{n, s}$ as the evidence that $s \in T$, i.e., the underlying tree covers node $s$. Theorem \ref{thm:new-formula} indicates that more weights are assigned to $s_{n, l}$ than $s_{n, l-1}$, i.e., $\omega_{n, l} > \omega_{n, l-1}$, if $\lambda$ is smaller (i.e., node $s_{n, l-1}$ is more likely to branch and thus less likely to be a leaf node), $p^{e}_{n, s_{n, l}} - p^{e}_{n, s_{n, l-1}}$ is larger (i.e., $s_{n, l}$ has more evidence than $s_{n, l-1}$) and $\sum_{q \in \mathcal{A}} \ell^{w}_{n, q s_{n, l-1}} - \ell^{w}_{n, s_{n, l}}$ is larger (i.e., $s_{n, l}$'s siblings have more evidence to explain the data and thus $s_{n, l-1}$ is less likely to be a leaf node). The indicator function $\mathbb{I}_{l = D}$ is due to the maximum depth constraint on the branching process. Nodes at level $l = D$ automatically stop the branching, i.e., the branching-stopping probability is $1$ for such nodes. 

We remark that the form in (\ref{eqn:next-token}) is similar in spirit to a formula in \cite{matsushima1994bayes}. However, $\omega_{n, l}$ was not viewed as a computable quantity in that early work, but only as a conditional distribution value for their subsequent theoretical analysis. 

\begin{proof}[Proof of Theorem \ref{thm:new-formula}]
Recall $s_{i, l} = (x_{i-l+1}, \ldots, x_{i})$ is the suffix at position $i$ of length $l$. %Note that when $i = n$ and it is clear from the context, we omit $n$ and write $s_{n, l}$ by $s_{n,l}$. 
We omit $D$ by writing $\mathcal{T} = \mathcal{T}(D)$ when $D$ is clear from the context. Define partition $\{ \mathcal{T}_{s_{n,l}} \}_{0 \leq l \leq D}$, that $\mathcal{T}_s = \{T \in \mathcal{T}: s \in \mathcal{L}(T)\}$ is the set of trees containing leaf $s$. The predicted probability can then be computed as
\begin{align}
\vec{P}_{\pi_{\text{CTW}}}(x_{n+1} | x_{1-D}^n) & = \sum_{T \in \mathcal{T}} \int p(x_{n+1} | T, \{p_s\} , x_{1-D}^n ) \pi_{\text{CTW}}(T, \{p_s\} | x_{1-D}^n ) {\Big (} \prod_{s \in \mathcal{L}(T)} \mathrm{d} p_s {\Big )} \notag \\
& = \sum_{l = 0, \ldots, D}~ \sum_{T \in \mathcal{T}_{s_{n,l}}} \int p_{s_{n,l}}(x_{n+1}) \pi_{\text{CTW}}(T, \{p_s\} | x_{1-D}^n ) {\Big (} \prod_{s \in \mathcal{L}(T)} \mathrm{d} p_s {\Big )} \notag \\
& = \sum_{l = 0, \ldots, D}~ \sum_{T \in \mathcal{T}_{s_{n,l}}} \int p_{s_{n,l}}(x_{n+1}) \pi_{D}(T | x_{1-D}^n) \pi_{p}( p_{s_l} | T, x_{1-D}^n) \mathrm{d} p_{s_l} \notag \\
& = \sum_{l = 0, \ldots, D}~ \sum_{T \in \mathcal{T}_{s_{n,l}}} \pi_D( T | x_{1-D}^n ) \int p_{s_{n,l}}(x_{n+1}) \pi_{p}( p_{s_l} | T, x_{1-D}^n ) \mathrm{d} p_{s_l} \notag \\
& = \sum_{l = 0, \ldots, D}~ \sum_{T \in \mathcal{T}_{s_{n,l}}} \pi_D( T | x_{1-D}^n ) \int p_{s_{n,l}}(x_{n+1}) \pi_{p}( p_{s_l} | \mathcal{T}_{s_{n,l}}, x_{1-D}^n ) \mathrm{d} p_{s_l} \notag \\
& = \sum_{l = 0, \ldots, D}~ {\Big (} \sum_{T \in \mathcal{T}_{s_{n,l}}} \pi_D( T | x_{1-D}^n) {\Big )} \left( \int p_{s_{n,l}}(x_{n+1}) \pi_{p}( p_{s_l} | \mathcal{T}_{s_{n,l}}, x_{1-D}^n) \mathrm{d} p_{s_l} \right) \notag \\
& = \sum_{l = 0, \ldots, D}~  \omega_{n, l} \cdot \vec{p}_{n, s_{n, l}}(x_{n+1}),\label{eqn:app-next-token}
\end{align}
where the last equality is by the definition that
\begin{align}
    \omega_{n, l} = \sum_{T \in \mathcal{T}_{s_{n,l}}} \pi_D( T | x_{1-D}^n), \label{eqn:omega}
\end{align}
and the optimal prediction probability given suffix $s_{n, l}$ is 
\begin{align}
    \vec{p}_{n, s_{n, l}}(a) = \frac{ \boldsymbol{\alpha}(a) + \vec{n}_{n,s_{n,l}}(a) }{ \sum_{q \in \mathcal{A}} (\boldsymbol{\alpha}(q) + \vec{n}_{n,s_{n,l}}(q)) }, \label{eqn:dirichlet-mean}
\end{align}
since for any $T \in \mathcal{T}_{s_{l}}$, the posterior of $p_{s}$ follows the Dirichlet distribution
\begin{align}
    \pi_{p}(p_{s_l} | T , x_{1-D}^n) = \text{Dir}(p_{s_l}; \boldsymbol{\alpha} + \vec{n}_{n, s_{n, l}}), 
\end{align}
with posterior mean $\mathbb{E}[p_{s_l} | T , x_{1-D}^n ] \in \Delta_{\mathcal{A}}$ and proportional to $\boldsymbol{\alpha} + \vec{n}_{n,s_{n,l}}$, which we had simply denoted as $\pi_{p}( p_{s_l} | \mathcal{T}_{s_{n,l}}, x_{1-D}^n)$.

Since the length of data $n$ is fixed and clear from the context, let $\underline{x} = x_{1-D}^n$ be the sequence, and we omit $n$ in the subscript of $p^{e}_{n, s}$, $p^w_{n, s}$ and $s_{n, l}$ for simplicity.

For any model $T \in \mathcal{T}(D)$, the posterior probability $\pi(T | \underline{x})$ is given by:
\begin{equation}
\pi_D(T | \underline{x} ) = \frac{\pi_D(T) P_\pi(\underline{x} | T)}{P_\pi(\underline{x})} = \frac{\pi_D(T) \prod_{s \in \mathcal{L}(T)} p^{e}_{s}}{p^{w}_{ () }},
\label{eq:mpost}
\end{equation}
where the denominator $P_\pi^*(\underline{x}) = p^{w}_{ ()}$ is the prior predictive likelihood computed by CTW given in Theorem~\ref{lem:predicteD-likelihood}, and the numerator is by $P_\pi(\underline{x} | T) = \prod_{s \in \mathcal{L}(T)} p^{e}_{s}$ in \citep[Lemma 2.2]{kontoyiannis2022bayesian}. 
Since $\omega_l = \sum_{T \in \mathcal{T}_{s_l}} \pi(T | \underline{x})$ by definition, we have for any $l = 1, 2, \ldots, D$,
\begin{align}
    \frac{\omega_{l}}{\omega_{l-1}} = \frac{ \sum_{T' \in \mathcal{T}_{s_{l}}} \pi_{D}( T' | x) }{ \sum_{T \in \mathcal{T}_{s_{l-1}}} \pi_{D}( T | x) } =  \frac{ \sum_{T' \in \mathcal{T}_{s_{l}}} \pi_{D}( T' ) \prod_{s \in \mathcal{L}(T')} p^{e}_{s} }{ \sum_{T \in \mathcal{T}_{s_{l-1}}} \pi_{D}( T ) \prod_{s \in \mathcal{L}(T)} p^{e}_{s} }.
\end{align}

Note that tree in $\mathcal{T}_{s_{l}}$ and trees in $\mathcal{T}_{s_{l-1}}$ share similarities. For any $T \in \mathcal{T}_{s_{l-1}}$, let $\mathcal{T}_{s_{l}; T} = \{T' \in \mathcal{T}_{s_{l}}: \mathcal{L}(T) \subset \mathcal{L}(T') \cup \{s_{l-1}\} \}$ be the set of trees that differs from $T$ only at subtree $\text{sub}(T'; s_{l}):= \{\text{subtree of $T'$ with root at $s$}\}$.

Take any $l = 1,2,\ldots, D-1$. For any $T \in \mathcal{T}_{s_{l-1}}$ and $T' \in \mathcal{T}_{s_l; T}$. Based on the definition of $\pi_D = (1 - \lambda)^{( |\mathcal{L}(T)| - 1 ) / (A - 1)} \lambda^{ |\mathcal{L}(T)| - |\mathcal{L}_D(T)| }$, it is not hard to verify that 
\begin{align*}
    \frac{\pi_D(T')}{\pi_D(T)} & = \frac{ \pi_{D-l+1}(\text{sub}(T'; s_{l-1})) }{ \pi_{D-l+1}(\text{sub}(T; s_{l-1})) } \\
    & = \frac{ (1-\lambda) \pi_{D-l}(\text{sub}(T'; s_{l})) \prod_{s'_l \in \text{sib}(s_l)} \pi_{D-l}(\text{sub}(T'; s_{l}')) }{ \lambda } \\
    & = (1-\lambda) \prod_{s'_{l} \in \text{sib}(s_{l}) } \pi_{D-l}( \text{sub}(T'; s_{l}')),
\end{align*}
where $\text{sib}(s_{l+1}) = \{q s_{l}: q \in \mathcal{A} \text{ and } q s_{l} \not= s_{l+1} \}$ is set of siblings of $s_{l+1}$. We can interpret the ratio as follows. $T'$ and $T$ only differs at the $\text{sub}(T'; s_{l-1})$ and $\text{sub}(T; s_{l-1})$. Since $T'$ branches at node $s_{l-1}$, we thus have the numerator in the second equation, where $(1-\lambda)$ corresponds to the branching, and then compute for the subtrees. Note that $T$ stops branching at $s_{l-1}$ and $T'$ stops branching at $s_{l}$, then $\pi_{D-l+1}(\text{sub}(T; s_{l-1})) = \pi_{D-l}(\text{sub}(T'; s_{l})) = \lambda$ equals to the stopping probability. 

Given any suffix $s$ with $|s| \leq D$, it has been shown in \citep[Proof of Theorem 3.1]{kontoyiannis2022bayesian} that for any $l \leq D$, 
\begin{align}
    p^{w}_{ s} = \sum_{U \in \mathcal{T}(D-l)} \pi_{D-l}(U) \prod_{u \in \mathcal{L}(U)} p^{e}_{us},
\end{align}
where $\mathcal{T}(D-l)$ is the set of trees with maximum depth $D-l$ and $\pi_{D-l}$ is the prior for bounded branching process with maximum depth $D-l$. We thus have
\begin{align}
    & \frac{ \sum_{T' \in \mathcal{T}_{s_{l}; T}} \pi_D(T') \prod_{s \in \mathcal{L}(T') }  p^{e}_{ s} }{ \pi_D(T) \prod_{s \in \mathcal{L}(T)}  p^{e}_{ s} } = \frac{ \sum_{T' \in \mathcal{T}_{s_{l}; T}} \pi_D(T') \prod_{s \in \mathcal{L}(T') }  p^{e}_{ s} }{ \pi_D(T) \prod_{s \in \mathcal{L}(T)}  p^{e}_{ s} } \\
    & = \sum_{T' \in \mathcal{T}_{s_{l}; T}} \frac{\pi_D(T')}{\pi_D(T)} \frac{ \prod_{s \in \mathcal{L}(T') \setminus \mathcal{L}(T) }  p^{e}_{ s} }{  p^{e}_{ s_{l-1}} } \\
    & = \sum_{T' \in \mathcal{T}_{s_{l}; T}} \left( (1-\lambda) \prod_{s'_{l} \in \text{sib}(s_{l}) } \pi_{D-l}( \text{sub}(T'; s_{l}')) \right) \left( \frac{ p^{e}_{ s_{l}} \prod_{s'_{l} \in \text{sib}(T; s_{l}) } \prod_{s \in \mathcal{L}(\text{sub}(T'; s'_{l}))}  p^{e}_{ s} }{  p^{e}_{ s_{l-1}} }\right) \\
    & = (1-\lambda) \frac{ p^{e}_{ s_{l}} }{  p^{e}_{ s_{l-1}} } \sum_{T' \in \mathcal{T}_{s_{l}; T}} \left( \prod_{s'_{l} \in \text{sib}(s_{l}) } \pi_{D-l}( \text{sub}(T'; s_{l}')) \right)  \left( \prod_{s'_{l} \in \text{sib}(T; s_{l}) } \prod_{s \in \mathcal{L}(\text{sub}(T'; s'_{l}))}  p^{e}_{ s} \right) \\
    & = (1-\lambda) \frac{ p^{e}_{ s_{l}} }{  p^{e}_{ s_{l-1}} } \sum_{T' \in \mathcal{T}_{s_{l}; T}} \left( \prod_{s'_{l} \in \text{sib}(s_{l}) } \pi_{D-l}( \text{sub}(T'; s_{l}')) \prod_{s \in \mathcal{L}(\text{sub}(T'; s'_{l}))}  p^{e}_{ s} \right) \\
    & = (1-\lambda) \frac{ p^{e}_{ s_{l}} }{  p^{e}_{ s_{l-1}} } \prod_{s'_{l} \in \text{sib}(s_{l})} \left( \sum_{U \in \mathcal{T}(D-l)} \pi_{D-l}(U) \prod_{u \in \mathcal{L}(U)} p^{e}_{ u s'_{l}} \right) \\
    & = \frac{(1-\lambda) p^{e}_{s_{l}} \prod_{a \not= s_{l} \setminus s_{l-1}} p^{w}_{ as_{l-1}}}{ p^{e}_{s_{l-1}}}. 
\end{align}

Similarly, for any $T \in \mathcal{T}_{s_{D-1}}$ and $T' \in \mathcal{T}_{s_D; T}$, $\frac{\pi_D(T')}{\pi_D(T)} = \frac{1-\lambda}{\lambda}$, we can derive
\begin{align}
    \frac{\omega_{D}}{\omega_{D-1}} = \frac{ (1-\lambda) p^{e}_{s_{D}} \prod_{a \not= s_{D} \setminus s_i} p^{w}_{ as_{D-1}}}{ \lambda p^{e}_{s_{D-1}}},
\end{align}
in the same manner. The proof can then be concluded by taking the logarithm on both sides. 
\end{proof}

\subsection{Construction of Transformer for CTW} \label{app:construction-ctw}

To make the presentation clear, in the following, we separate the layers by their functionality and present them separately. 
Recall that
\begin{align*}
\vec{a}_{i}^{(\ell)}=\text{MHA}\left(\vec{h_i},\vec{H};\{W_{O,m}^{(\ell)},W_{Q,m}^{(\ell)},W_{K,m}^{(\ell)},W_{V,m}^{(\ell)}\}_{m=1}^{M^{(\ell)}}\right)\triangleq W_O^{(\ell)}\left[\vec{b}_{1,i}^{(\ell)};\vec{b}_{2,i}^{(\ell)};\ldots;\vec{b}_{M^{(\ell)},i}^{(\ell)}\right],
\end{align*}
where $\{W_{Q,m}^{(\ell)},W_{K,m}^{(\ell)},W_{V,m}^{(\ell)}\}_{m=1}^{M^{(\ell)}}$ are the $E^{(\ell)} \times E$ query matrices, key matrices, and value matrices and $W_{O}^{(\ell)}$ is the $E \times M^{(\ell)}E^{(\ell)}$ output mapping matrix. For simplicity of presentation, we take $E^{\ell} = E$ and $W_O^{\ell} = [\vec{I}; \vec{I}; \ldots; \vec{I}]$. It is not hard to see the following constructions can be applied to much smaller $E^{(\ell)}$ while taking $W_{O}$ as a permutation matrix.

\subsubsection{Finite-memory context-extension layer} \label{app:extension} 

We begin with the first layer, which is referred to as a finite-memory context-extension layer. 

\begin{theorem}[Restate Theorem \ref{thm:extension}]
There is an $M$-headed transformer layer that can perform finite-memory context extension, defined by the following output, with the initial one-hot embedded input $\vec{H}^{(1)}$:
\begin{align}
\vec{h}^{(2)}_i = (\vec{s}_{i, M+1}; \vec{0}; \vec{pos}_i) = (\vec{x}_i;\vec{x}_{i-1};\ldots;\vec{x}_{i-M}; \vec{0}; \vec{pos}_i), \label{eqn:statistics-also}
\end{align}
where $\vec{s}_{i, M+1} = (\vec{x}_i;\ldots;\vec{x}_{i-M})$ is the vector version of the $M$-length suffix $s_{i, M+1} = x_{i - M}^{i}$.
\end{theorem}
\begin{proof}[Proof of Theorem \ref{thm:extension}] 

The multi-head attention in the first layer is consisted of $M^{(1)} = M$ heads parameterized by $(W^{(1)}_{Q, m}, W^{(1)}_{K, m}, W^{(1)}_{V, m})_{m = 1,2,\ldots, M^{(1)}}$. Specifically, for $m = 1, 2, \ldots, M^{(1)}$,
\begin{align}
    W_{Q,m}^{(1)} = 
    \begin{pmatrix}
        \vec{0} & \text{Rot}(m) \\
        \vec{0} & \vec{0}
    \end{pmatrix}, \quad
    W_{K, m}^{(1)} = 
    \begin{pmatrix}
        \vec{0} & c \vec{I}^{2 \times 2} \\
        \vec{0} & \vec{0}
    \end{pmatrix}, \quad
    W_{V, m}^{(1)} =
    \begin{pmatrix}
        \vec{0}^{mA \times A} & \vec{0} \\
        \vec{I}^{A \times A} & \vec{0} \\
        \vec{0} & \vec{0}
    \end{pmatrix},
\end{align}
where $\text{Rot}(m) = \begin{pmatrix}
\cos(m\pi/N) & \sin(m\pi/N) \\  
-\sin(m\pi/N) & \cos(m\pi/N) 
\end{pmatrix}$ is a rotation matrix that rotates clockwise by an angle of $m\pi/C$, and $c \in \mathbb{R}_+$ is a temperature factor. The query, key, and value after the mapping are
\begin{align}
    W^{(1)}_{Q, m} \vec{h}_n^{(1)} = 
    \begin{pmatrix}
        \vec{pos}_{n - m} \\
        \vec{0}
    \end{pmatrix}, \quad
    W^{(1)}_{K, m} \vec{h}_i^{(1)} = c 
    \begin{pmatrix}
        \vec{pos}_{i} \\
        \vec{0}
    \end{pmatrix}, \quad
    W^{(1)}_{V, m} \vec{h}_i^{(1)} = 
    \begin{pmatrix}
        \vec{0}^{mA \times 1} \\
        \vec{x}_i \\
        \vec{0}
    \end{pmatrix}.
\end{align} 
Take $c=\infty$ or sufficiently large. It is seen that the $m$-th head essentially copies the $m$-th earlier symbol to stack at the $(m+1)$-th position below the original symbol $\vec{x}_i$. Together with the residual link, the attention layer gives exactly the $\vec{h}^{(2)}_i$ shown in (\ref{eq:finite-memory-also}) while the FF layer in this layer can be set to all zeros. 
\begin{align}
\vec{h}^{(2)}_i=(\vec{x}_i;\vec{x}_{i-1};\vec{x}_{i-2};\vec{x}_{i-M^{(1)}}; \vec{0}; \vec{pos}_i) = (\vec{s}_{i, M^{(1)}+1}; \vec{0}; \vec{pos}_i), \label{eq:finite-memory-also}
\end{align}
where $\vec{s}_{i, l} = (\vec{x}_i;\vec{x}_{i-1};\cdots; \vec{x}_{i-l+1})$ is the one-hot embedded version of suffix $s_{i,l} = (x_{i - l +1}, \ldots, x_{i - 1}, x_i)$. 
\end{proof}

\subsubsection{Statistics collection layer}

\begin{theorem}
[Restate Theorem \ref{thm:statistics}]
There is an $M'$-head attention layer, where $M'\leq M+1$, that can perform statistics collection, defined by the following output, with $\vec{H}^{(2)}$ in (\ref{eq:finite-memory}) as its input:
\begin{align}
    \vec{a}^{(2)}_i=(\vec{s}_{i, M+1}; \vec{g}_{i, M'}; \vec{g}^{\leftarrow}_{i-1, M'} ; \vec{0}; \vec{pos}_i), \label{eqn:app-statistics}
\end{align}
where $\vec{g}_{i, M'} := (\vec{g}_{i, s_{i, 0}}; \ldots;\vec{g}_{i, s_{i, M'-1}})$ and $\vec{g}^{\leftarrow}_{i-1, M'} = (\vec{g}^{\leftarrow}_{i-1, s_{i, 0}};\ldots;\vec{g}^{\leftarrow}_{i-1, s_{i, M'-1}})$.
\end{theorem}
\begin{proof}[Proof of Theorem \ref{thm:statistics}]
To make the proof self-contained, we first recall some key notations. The second layer is referred to as the statistics collection layer, which uses a sequence of vectors $\vec{h}^{(2)}_i$, $i=1,2,\ldots,N$, defined in (\ref{eq:finite-memory}) as its input, restated as follows. 
\begin{align}
\vec{h}^{(2)}_i = (\vec{s}_{i, M+1}; \vec{0}; \vec{pos}_i),
\end{align}
where $\vec{s}_{i, M+1} = (\vec{x}_i;\ldots;\vec{x}_{i-M})$. To rigorously specify the function of this layer, recall the definition of the $k$-gram statistics vector $\vec{g}_{i, s}$, which in plain words, is the empirical probability distribution of the next token associated with the suffix $s$ for a sequence $x_{1}^{i}$. Mathematically, for a suffix $s$ whose length is $k-1$ and the current position $i$, 
\begin{align}
    \vec{g}_{i, s}(a) = \frac{\vec{n}_{i, s}(a)}{ \sum_{q \in \mathcal{A}} \vec{n}_{i, s}(q)}\quad \forall a \in \mathcal{A}, \label{eqn:forward-statistics}
\end{align}
where $\vec{n}_{i, s}$ is the counting vector defined in (\ref{eqn:counting-vector}). 

The $k$-gram backward statistics vector $\vec{g}_{i-1, s}^{\leftarrow}$ is defined similarly, which is the empirical probability distribution of the previous token associated with the suffix $s$ for data $x_{1}^{i-1}$, and mathematically
\begin{align}
    \vec{g}_{i-1, s}^{\leftarrow}(a) = \frac{ \sum_{q \in \mathcal{A}} \vec{n}_{i, as}(q) }{ \sum_{q \in \mathcal{A}} \vec{n}_{i, s}(q) }\quad \forall a \in \mathcal{A}, \label{eqn:backward-statistics}
\end{align}
where $\sum_{q \in \mathcal{A}} \vec{n}_{i, s}(q)$ is the number of appears of the sub-string $s$ in the sequence $x_{1}^{i-1}$. 

The multi-head attention in the second layer is consisted of $M^{(2)} = M' \leq M^{(1)} + 1 = M + 1$ heads parameterized by $(W^{(2)}_{Q, m}, W^{(2)}_{K, m}, W^{(2)}_{V, m})_{m = 0, 1,2,\ldots, M^{(2)}-1 }$. Specifically, for $m = 1, 2, \ldots, M^{(2)} - 1 $,
\begin{align}
    W_{Q, m}^{(2)} &= 
    \begin{pmatrix}
        \vec{I}^{(m-1)A \times (m-1)A} & \vec{0} \\
        \vec{0} & \vec{0}
    \end{pmatrix},~
    W_{K, m}^{(2)} = 
    \begin{pmatrix}
        \vec{0}^{(m-1)A \times A} & c \vec{I}^{(m-1)A \times (m-1)A} & \vec{0} \\
        \vec{0} & \vec{0} & \vec{0}
    \end{pmatrix},\\ ~
    W_{V, m}^{(2)} &= 
    \begin{pmatrix}
        \vec{0}^{ (M^{(1)}+m)A \times A} & \vec{0} \\
        \vec{I}^{A \times A} & \vec{0} \\
        \vec{0}^{ (M^{(2)} -1) A \times A} & \vec{0} \\
        \vec{0}^{A \times A} & [\vec{0}^{A \times (m-1) A} , \vec{I}^{A \times A}, \vec{0}] \\
        \vec{0} & \vec{0}
    \end{pmatrix}.
\end{align}
The corresponding query, key, and value vectors after the mapping are
\begin{align*}
    W^{(2)}_{Q, m} \vec{h}_n^{(2)} & = 
    \begin{pmatrix}
        \vec{s}_{n, m-1} \\
        \vec{0}
    \end{pmatrix}, \quad
    W^{(2)}_{K, m} \vec{h}_i^{(2)} = c 
    \begin{pmatrix}
        \vec{s}_{i-1,m-1} \\
        \vec{0}
    \end{pmatrix}, \quad 
    W^{(2)}_{V, m} \vec{h}_i^{(2)} & = 
    \begin{pmatrix}
        \vec{0}^{(M^{(1)} + m) A \times 1} \\
        \vec{x}_i \\
        \vec{0}^{ (M^{(2)} - 1 ) A \times 1} \\
        \vec{x}_{i - m} \\
        \vec{0}
    \end{pmatrix}.
\end{align*}
For $m = M^{(2)}$, $W_{Q, m}^{(2)}, W_{K, m}^{(2)}$ are of the same structure, while $W^{(2)}_{V, m}$ does not contains that $\vec{I}^{A \times A}$ in that $[\vec{0}^{A \times (m-1) A} , \vec{I}^{A \times A}, \vec{0}]$ block, and thus $W^{(1)}_{V, m} \vec{h}_i^{(1)}$ does not have $\vec{x}_{i - m}$. 

It is not hard to see that taking $c \rightarrow \infty$ gives
\begin{align}
(\vec{s}_{i, M^{(1)}+1}; \vec{g}_{i, M^{(2)}-1}; \vec{g}^{\leftarrow}_{i-1, M^{(2)}-1} ; \vec{0}; \vec{pos}_i) = [ \text{MHA}(\vec{H}^{(2)}) + \vec{H}^{(2)} ]_i,
\end{align}
where 
\begin{align*}
    \vec{g}_{i, M'} & = (\vec{g}_{i, s_{i, 0}}; \ldots;\vec{g}_{i, s_{i, M'-1}}) \\ \vec{g}^{\leftarrow}_{i-1, M'} & = (\vec{g}^{\leftarrow}_{i-1, s_{i, 0}};\ldots;\vec{g}^{\leftarrow}_{i-1, s_{i, M'-1}}).
\end{align*}
\end{proof}

Note that the counting vector can be obtained via 
\begin{align}
    \vec{n}_{i, s_{i, l}}(a) & = \frac{\vec{n}_{i, s_{i, l}}(a)}{ \sum_{q \in \mathcal{A}} \vec{n}_{i, s_{i, l}}(q) } \frac{ \sum_{q \in \mathcal{A}} \vec{n}_{i, s_{i, l}}(q)}{ \sum_{q \in \mathcal{A}} \vec{n}_{i, s_{i, l-1}}(q) } \cdots \frac{ \sum_{q \in \mathcal{A}} \vec{n}_{i, s_{i, 1}}(q)}{ \sum_{q \in \mathcal{A}} \vec{n}_{i, s_{i, 0}}(q) } \left(\sum_{q \in \mathcal{A}} \vec{n}_{i, s_{i, 0}}(q) \right) \\
    & = \vec{g}_{i, s_{i, l}}(a) \left( \prod_{j = 0}^{l - 1}  \vec{g}_{i - 1, s_{i, j }}^{\leftarrow}(x_{i - j } ) \right) \cdot i,
\end{align}
by the information contained in vector $(\vec{s}_{i, M^{(1)}+1}; \vec{g}_{i, M^{(2)}-1}; \vec{g}^{\leftarrow}_{i-1, M^{(2)}-1} ; \vec{0}; \vec{pos}_i)$. 

Since $p_{i, s_{i, l}}^e$ and $\vec{p}_{i, s_{i, l}}$ in \eqref{eqn:dirichlet-mean} are simple functions of $\vec{n}_{i, s_{i, l}}$, they can be approximated by a sufficiently wide FF layer, yielding the following output
\begin{align}
    \vec{h}^{(3)}_i=(\vec{s}_{i, M^{(1)}+1}; \vec{p}_{i, D} ; \vec{l}^{e}_{i, D}; \ln(p^{w}_{i, s_{i, D}}) ; \vec{0}; \vec{pos}_i),
\end{align}
where $\vec{l}^{e}_{i, D}$ contains the logarithm of $p^e$ along the path from root $()$ to $(x_{i-d+1}, \ldots, x_{i})$, and $\vec{p}_{i, D}$ stacks the optimal prediction given suffices $s_{i, 0}, \ldots, s_{i, D}$, i.e., 
\begin{align}
    \vec{l}^{e}_{i, D} & = (\ell^{e}_{i, s_{i, 0}}; \ell^{e}_{i, s_{i, 1}}; \ldots; \ell^{e}_{i, s_{i, D}} ) = (\ln(p^{e}_{i, s_{i, 0}}); \ln(p^{e}_{i, s_{i, 1}}); \ldots; \ln(p^{e}_{i, s_{i, D}}) ), \\
    \vec{p}_{i, D} & = (\vec{p}_{i, s_{i, 0}}; \vec{p}_{i, s_{i, 1}}; \ldots; \vec{p}_{i, s_{i, D}}),
\end{align}
and $\ln(p^{w}_{i, s_{i, D}}) = \ln(p^{e}_{i, s_{i, D}})$ with suffix $|s_{i, D}| = D$. These quantities can be extracted, since they are functions of the statistics collected from $\vec{a}_i^{(2)}$. 

This functional layer essentially collects $k$-gram statistics for various lengths of $k=1,2,\ldots,M^{(2)}$ via multi-head attention and then processes the statistics for the follow-up optimal scheme.

\subsubsection{Inductive CTW layer}

Recall the inputs of the inductive CTW layers at $\ell=3,4,\ldots,3+D$ (layer-$(3+D)$ is a fictitious layer), which are also the outputs of layers $2,3,\ldots,2+D$, should be
\begin{align}
    \vec{h}^{(\ell)}_i=(\vec{s}_{i, M^{(1)}+1}; \vec{p}_{i, D} ; \vec{l}^{e}_{i, D}; \delta_{i, D}; \delta_{i, D-1} ; \ldots ; \delta_{i, D - \ell + 4} ; \ell^{w}_{i, s_{i, D+3-\ell}} ; \vec{0}; \vec{pos}_i),
    \label{eqn:h-ell-induction-also}
\end{align}
for $\ell = 3, 4, \ldots, 3 + D$, where $\delta_{i, l} := \ln(\omega_{i, l}) - \ln(\omega_{i, l-1})$ for $l = D, D-1, \ldots, 1$ are the weight difference, and we take $M^{(1)} = D$.

\begin{theorem}[Restatement of Theorem \ref{thm:induction}]
There exists a $A$-head transformer layer that can perform the induction: Takes $\vec{H}^{(\ell)}$ in (\ref{eqn:h-ell-induction-also}) as input and outputs $\vec{H}^{(\ell+1)}$. And the final readout layer taking $\vec{H}^{(D+2)}$ as input can output the $A$-dimensional Bayesian optimal next token prediction vector
$
P_{\pi_{\text{CTW}}}(\cdot | x_{1-D}^n) = \sum_{l = 0, \ldots, D}~  \omega_{n, l} \vec{p}_{n, s_{n, l}}.
$
\end{theorem}
\begin{proof}[Proof of Theorem \ref{thm:induction}]
For any fixed $\ell = 3, 4, \ldots, 2 + D$, we specify the construction for the $\ell$-th transformer layer. It contains $A$ heads and for each $m = 1, 2, \ldots, A$, the $Q, K, V$ matrices are
\begin{align*}
    W_{Q, m}^{(\ell)} &= 
    \begin{pmatrix}
        \vec{I}^{(D + 1 - \ell)A \times (D + 1 - \ell)A} & \vec{0} \\
        \vec{0} & [ \vec{e}_m, \vec{0}^{A \times 2}] \\
        \vec{0} & \vec{0} \\
        \vec{0} & \vec{I}^{2 \times 2}
    \end{pmatrix},~
    W_{K, m}^{(\ell)} = 
    \begin{pmatrix}
        c \vec{I}^{(D + 2 - \ell)A \times (D + 2 - \ell)A} & \vec{0} \\
        \vec{0} & \vec{0} \\
        \vec{0} & c\vec{I}^{2 \times 2}
    \end{pmatrix},
    \\ 
    W_{V, m}^{(\ell)} & = 
    \begin{pmatrix}
        \vec{0}^{ (\text{place}_{\ell} + m) \times (\text{place}_{\ell} + m) } & \vec{0} \\
        [\vec{0}^{1 \times (\text{place}_{\ell} - 1)}, 1] & \vec{0} \\
        \vec{0} & \vec{0}
    \end{pmatrix},
\end{align*}
where $\vec{e}_m$ is the $A$-dimensional one-hot vector at position $m$, and $\text{place}_{\ell} = (M^{(1)} + D + 2)A + D + \ell - 1$ is the index of element $\ell^{w}_{i, s_{i, D + 3 -\ell}}$ in $\vec{h}_i^{(\ell)}$. 
The corresponding query, key, and value vectors after the mapping are
\begin{align*}
    W^{(\ell)}_{Q, m} \vec{h}_n^{(\ell)} & = 
    \begin{pmatrix}
        \vec{s}_{n, D + 1 - \ell} \\
        \vec{e}_m \\
        \vec{0} \\
        \vec{pos}_{n}
    \end{pmatrix}, \quad
    W^{(\ell)}_{K, m} \vec{h}_i^{(\ell)} = c 
    \begin{pmatrix}
        \vec{s}_{i,D + 2 - \ell} \\
        \vec{0} \\
        \vec{pos}_{i}
    \end{pmatrix}, \quad 
    W^{(\ell)}_{V, m} \vec{h}_i^{(\ell)} = 
    \begin{pmatrix}
        \vec{0}^{ (\text{place}_{\ell} + m) \times 1} \\
        \ell^{w}_{i, s_{i, D + 3 -\ell}} \\
        \vec{0}
    \end{pmatrix}.
\end{align*}
At position $n$, the query of $m$-th head will select the latest (due to positional embedding) position with suffix $[\vec{s}_{n, D + 1 - \ell}; \vec{e}_{m} ]$, and append its $\ell^{w}$ at the end. It is not hard to see that taking $c \rightarrow \infty$ gives 
\begin{align*}
    & \vec{a}_i^{(\ell)} =  [ \text{MHA}(\vec{H}^{(2)}) + \vec{H}^{(2)} ]_i \\
    & = (\vec{s}_{i, D +1}; \vec{p}_{i, D} ; \vec{l}^{e}_{i, D}; \delta_{i, D}; \delta_{i, D-1} ; \ldots ; \delta_{i, D + 4 - \ell} ; \ell^{w}_{i, s_{i, D+3-\ell}} ; [\ell^{w}_{i, q s_{i, D+2-\ell}}]_{q \in \mathcal{A}} ;\vec{0}; \vec{pos}_i)
\end{align*}

Recall $\ln(\omega_{n, l}) - \ln(\omega_{n, l-1}) = \ln(1-\lambda) - \mathbb{I}_{l = D}\ln(\lambda) + \ell^{e}_{n, s_{n, l}} - \ell^{e}_{n, s_{n, l-1}} + \sum_{q \in \mathcal{A}} \ell^{w}_{n, q s_{n, l-1}} - \ell^{w}_{n, s_{n, l}}$ by Theorem \ref{thm:main-new-formula}. $\delta_{i, D + 3-\ell} = \ln(\omega_{i, D + 3 - \ell}) - \ln(\omega_{i, D + 2 - \ell})$ can be computed by $\vec{a}_i^{(\ell)}$ and thus $\vec{h}_{i}^{(\ell + 1)}$ can be conveniently computed via the FF layer following the $\ell$-th multi-head attention layer. 

The final readout layer approximates an $A$-dimensional vector
\begin{align}
P_{\pi_{\text{CTW}}}(\cdot | x_{1-D}^n) = \sum_{l = 0, \ldots, D}~  \omega_{n, l} \cdot \vec{p}_{n, s_{n, l}}( \cdot ),
\end{align}
by an FF layer taking input 
\begin{align}
    \vec{h}^{(D+3)}_n =(\vec{s}_{n, M^{(1)}+1}; \vec{p}_{n, D} ; \vec{l}^{e}_{n, D}; \delta_{n, D}; \ldots ; \delta_{n, 1}; \vec{0}; \vec{pos}_i).
\end{align}
The proof is now complete.
\end{proof}

\end{document}